\documentclass[journal=jacsat,manuscript=article]{achemso}

\usepackage{chemformula} 
\usepackage[T1]{fontenc} 
\usepackage{subfigure}
\usepackage{cleveref}
\usepackage{multirow}

\usepackage[normalem]{ulem}
\author{Pei Zhang}
\affiliation{Computational Sciences and Engineering Division, Oak Ridge National Laboratory, Oak Ridge, TN, USA}
\email{zhangp1@ornl.gov}
\phone{+1 865 341 0368}
\author{Logan Kearney}
\affiliation{Chemical Sciences Division, Oak Ridge National Laboratory, Oak Ridge, TN, USA}
\author{ Debsindhu Bhowmik}
\affiliation{Computational Sciences and Engineering Division, Oak Ridge National Laboratory, Oak Ridge, TN, USA}
\author{Zachary Fox}
\affiliation{Computational Sciences and Engineering Division, Oak Ridge National Laboratory, Oak Ridge, TN, USA}
\author{Amit~K.~Naskar}
\affiliation{Chemical Sciences Division, Oak Ridge National Laboratory, Oak Ridge, TN, USA}
\author{John Gounley}
\affiliation{Computational Sciences and Engineering Division, Oak Ridge National Laboratory, Oak Ridge, TN, USA}

\title[An \textsf{achemso} demo]
  {Transferring a molecular foundation model for polymer property predictions 
  }
\abbreviations{IR,NMR,UV}

\begin{document}

\begin{tocentry}

            \includegraphics[width=1.0\linewidth]{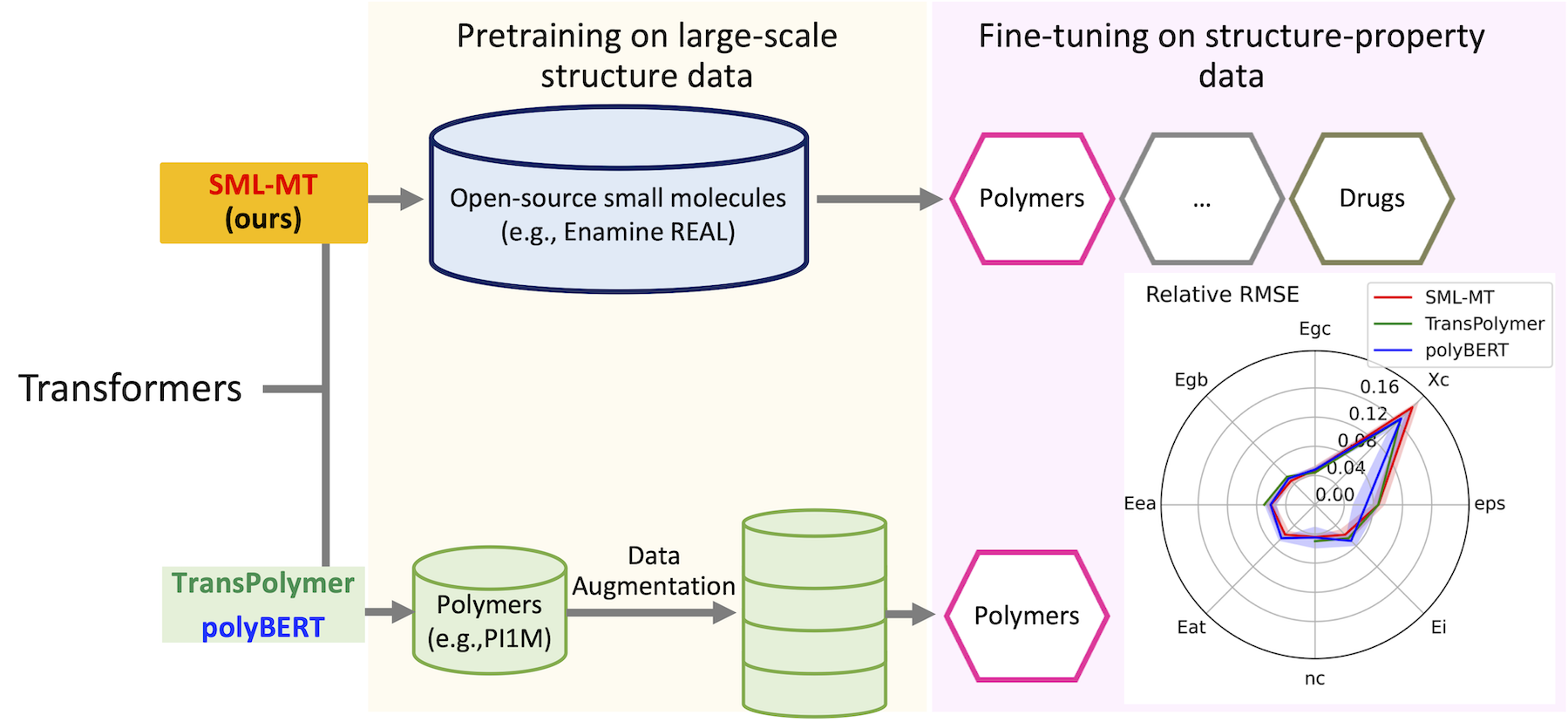}





\end{tocentry}

\begin{abstract}
Transformer-based large language models have remarkable potential to accelerate design optimization for applications such as drug development and materials discovery. Self-supervised pretraining of transformer models requires large-scale datasets, which are often sparsely populated in topical areas such as polymer science. State-of-the-art approaches for polymers conduct data augmentation to generate additional samples but unavoidably incurs extra computational costs. In contrast, large-scale open-source datasets are available for small molecules and provide a potential solution to data scarcity through transfer learning. In this work, we show that using transformers pretrained on small molecules and fine-tuned on polymer properties achieve comparable accuracy to those trained on augmented polymer datasets for a series of benchmark prediction tasks.

\end{abstract}

\section{Introduction}

Polymer informatics is a rapidly growing field that leverages data-centric machine learning (ML) approaches to discover and design new polymers \cite{audus2017polymer, cencer_machine_2022, chen_polymer_2021}. For efficient exploration of the vast chemical space, fast ML surrogates \cite{tao_benchmarking_2021, jablonka_bias_2021, kuenneth2022bioplastic,nagasawa_computer-aided_2018,kuenneth_copolymer_2021, sattari_data-driven_2021, ma_evaluating_2019, jha_impact_2019, tao_machine_2021,doan_tran_machine-learning_2020, kamal_novel_2021}
are often developed for polymer structure-property relationships. Although these surrogates differ in structure representations (e.g., repeat units, polymer chains), feature representations (e.g., string representation, molecular graph, chemical descriptors), and ML algorithms (e.g., random forest, Gaussian process regression, deep neural network [DNN]), they focus on conventional small-scale ML models in the supervised learning paradigm and suffer from inadequately labeled (i.e., structure-property) data issues.  
Recently, transformer-based large language models (LLMs) TransPolymer \cite{xu2023transpolymer} and polyBERT \cite{kuenneth_polybert_2023} have achieved state-of-the-art accuracy for prediction of polymer properties. 
However, training such LLMs requires data at scale, which is not readily publicly available for polymers.
In this study, we evaluate whether the key limitation for developing language models for polymers---the lack of large, publicly available polymer datasets---can be mitigated with a foundation model approach.

Transformers, which were originally developed for natural language processing \cite{vaswani_attention_2017}, have been applied to guide design in multiple scientific applications (e.g., predicting function from protein sequences\cite{kotsiliti2022novo,meier2021language,madani2023large, ferruz2022protgpt2, ferruz2022controllable,brandes2022proteinbert} and generating small molecules with desired properties \cite{chithrananda2020chemberta,Blanchard2022, blanchard2023adaptive,ross2022large,ahmad2022chemberta}). Two factors have facilitated the development of LLMs for these applications. First, these domains have text-based sequence representations (e.g., FASTA \cite{lipman1985rapid} and the simplified molecular-input line-entry system [SMILES]) that are easily amenable to representation as natural language text. Second, these applications have massive, previously compiled datasets (e.g., Enamine REAL~\cite{shivanyuk2007enamine}, PubChem \cite{kim2021pubchem}, and ZINC \cite{irwin2005zinc} for molecules and Uniprot \cite{uniprot2019uniprot} for proteins) that have enabled the training of these LLMs.

The typical training process involves two stages. First, \textbf{pretraining} is primarily self-supervised on large-scale structure data (e.g., SMILES), and the model learns to encode the generic structure information. Second, \textbf{fine-tuning} is supervised on smaller task-specific labeled data (i.e., properties), and the pretrained model learns to predict properties from structures. 
For example, when ChemBERTa-2 \cite{ahmad2022chemberta} was pretrained on 77 million compounds with a SMILES representation from PubChem and fined-tuned on tasks (with 1,500$\sim$8,000 labeled samples) from MoleculeNet and when ProteinBERT \cite{brandes2022proteinbert} was pretrained on about 106 million proteins from Uniprot UniRef90 and fine-tuned on nine benchmarks with a training size varying from 2,700 to 53,700, they both achieved the best accuracy for most benchmark tasks. 
These advancements in LLMs for molecules and proteins are attributed to the high representation capacity of the transformer architecture and, equally important, the diverse large-scale datasets available for pretraining such models.

To improve learning for polymer-property prediction in the presence of limited fine-tuning data, the potential value of large-scale pretraining is clear. Open-source datasets for polymer properties from DFT simulations \cite{kuenneth2021polymer} are usually small with a few hundred or a few thousand samples but are nonetheless larger than datasets of experimental measurements. However, the data scarcity challenge for polymers extends to large-scale structure data for self-supervised pretraining as well. Large polymer databases such as PolyInfo \cite{otsuka2011polyinfo} and Polymer Genome \cite{kim2018polymer} do not provide access to curated downloadable datasets. On the other hand, the accessible PI1M \cite{ma_pi1m_2020} database does contain about 1 million polymer structures, but it has been compiled by using a generative AI approach based on a recurrent neural network trained on a much smaller dataset. Not only is PI1M still orders of magnitude smaller than datasets such as Enamine REAL, but the manner of its generation may limit the diverse representations necessary for language model pretraining. 
Still, TransPolymer \cite{xu2023transpolymer} obtained state-of-the-art results for property prediction by leveraging pretraining on about 5 million noncanonical SMILES entries augmented from PI1M. 
Similarly, polyBERT \cite{kuenneth_polybert_2023} created its 100~million--sample pretraining dataset by fragment-based data augmentation on about 13,800 
synthesized polymers.

In this work, we evaluate whether pretraining a transformer on a large-scale small molecule dataset such as Enamine REAL can lead to a model that is similarly effective to TransPolymer or polyBERT. Our hypothesis is that the information being learned during pretraining, whether on datasets of polymers or small molecules, is primarily generic chemical language that serves as a suitable starting point for fine-tuning property prediction. To test this hypothesis, we pretrain BERT language models \cite{devlin2018bert} on PI1M and Enamine REAL. Along with a randomly initialized model that serves as a control, we evaluate the models on downstream tasks for predicting polymer properties after supervised fine-tuning on an open-source DFT dataset \cite{kuenneth2021polymer}. This labeled DFT dataset is much smaller than typically desired for deep learning, and this should emphasize the role of pretraining. In similar fashion to Kuenneth et al. \cite{kuenneth2021polymer, kuenneth_polybert_2023}, we employ a multitasking approach which predicts all properties simultaneously for better data efficiency. Our results show that the models pretrained on Enamine REAL achieve similar accuracy to the TransPolymer and polyBERT work, thereby suggesting that a model pretrained on a large-scale, accessible, and general molecular dataset could serve as a foundation model for applications across the molecular design landscape. Such an approach---pretraining on small molecular structures and fine-tuning for multitasking with multiple properties simultaneously (SML-MT)---would avoid the cost of generating and pretraining on large augmented polymer datasets. Moreover, the randomly initialized model learns quite effectively, and this suggests clear limits to the value of current pretraining procedures for molecular data.

\section{Core results}

We trained transformers for polymer properties following the standard two-step procedure: self-supervised pretraining on a large-scale structure dataset and supervised fine-tuning on small structure-property datasets (i.e., the DFT dataset \cite{kuenneth2021polymer} with 8 properties in Table \ref{tab:table-dft}).
The fine-tuned models are evaluated based on their accuracy when predicting properties for polymers unseen in training.
The evaluation metrics are relative root mean square error (RMSE) and R$^2$ with a 5-fold cross-validation, which are similar to those used for TransPolymer \citep{xu2023transpolymer} and polyBERT \citep{ kuenneth_polybert_2023}. 
The relative RMSE is defined as RMSE scaled by the difference between the maximum and minimum values of each property.

\begin{table}[h!]
  \begin{center}
    \caption{Open-source DFT dataset with 8 properties \cite{kuenneth2021polymer}}
    \label{tab:table-dft}
    \begin{tabular}{l|c|c|c|c} 
      \textbf{Property} & \textbf{Unit} & \textbf{Number of samples} & \textbf{Minimum} & \textbf{Maximum} \\
      \hline
      Atomization energy (Eat) &eV$\cdot$atom$^{-1}$ & 390 &$-$7.0&$-$5.0\\
      Crystallization tendency (Xc) & \% & 432 &0.10&100.0\\
      Bandgap (chain) (Egc) & eV & 3,380&0.02&10.0\\
      Bandgap (bulk) (Egb) & eV & 561&0.4&10.0\\
      Electron affinity (Eea) & eV & 368&0.4&5.0\\
      Ionization energy (Ei) & eV & 370&4.0&10.0\\
      Refractive index (nc) & 1 & 382&1.0&3.0\\
      Dielectric constant (eps) & 1 & 382&3.0&9.0\\
      \hline
      In total & & 6,265&&\\
    \end{tabular}
  \end{center}
\end{table}

\begin{figure}[!ht]
\centering
\includegraphics[width=0.9\linewidth]{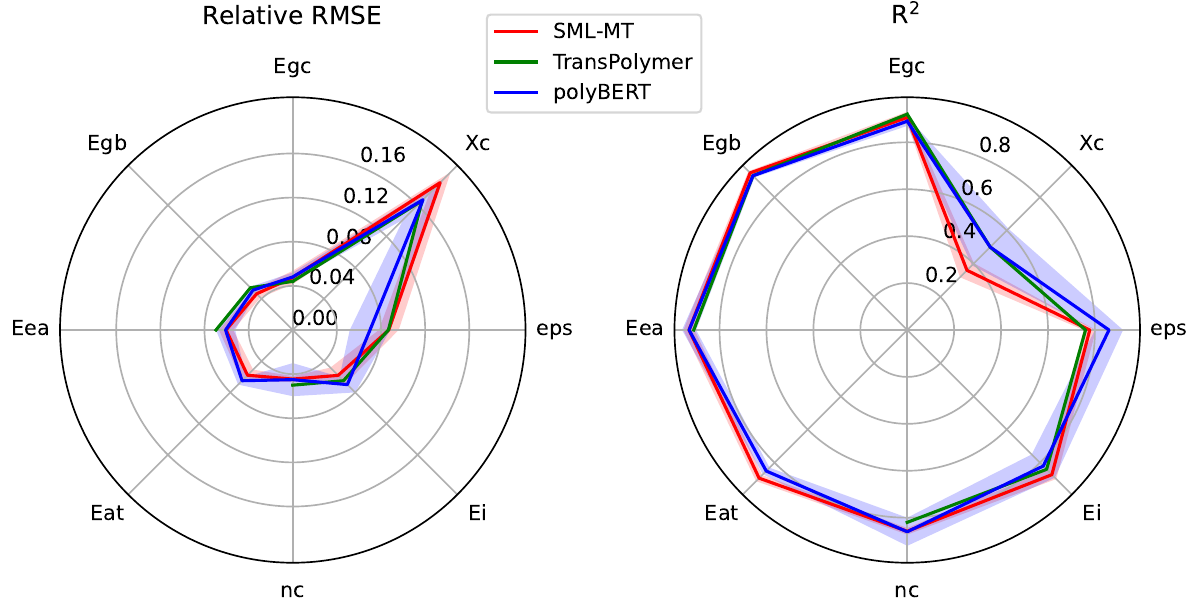}
\caption{Prediction performance metrics (relative RMSE and R$^2$) of SML-MT, TransPolymer, and polyBERT for 8 DFT properties. The lines represent the mean values and the shaded areas represent the standard deviation (STD) values of the test sets in the 5-fold cross-validation. Performance data for TransPolymer is reproduced from Xu et al.'s Table 4 \cite{xu2023transpolymer}, which has neither Eat nor STD values, and polyBERT is reproduced from Kuenneth et al.'s Fig. 5 and Supplementary Table S1 \cite{kuenneth_polybert_2023}. SML-MT, pretrained on small molecules, achieves similar accuracy to TransPolymer and polyBERT, pretrained on augmented polymer datasets. In particular, of 8 DFT properties, SML-MT has the best accuracy for 5 (i.e., Egb, Eea, Eat, nc, and Ei), comparable accuracy for 2 (i.e., Egc and eps), and poor accuracy for 1 (i.e., Xc). The poor prediction performance of all three models on Xc (i.e., crystallization tendency) is expected. The crystallization process highly depends on polymer structure descriptions across multiple length scales, and the current SMILES representation for only repeat units is inadequate.  
}
\label{fig:polar-R2RMSE}
\end{figure}

Figure~\ref{fig:polar-R2RMSE} shows a comparison of the prediction accuracy of our SML-MT model, which was pretrained on Enamine REAL and fine-tuned in the multitasking context, versus the accuracy of TransPolymer \cite{xu2023transpolymer} and polyBERT \cite{kuenneth_polybert_2023}. 
The three surrogates are trained differently.
\begin{itemize}
    \item TransPolymer \cite{xu2023transpolymer} is based on RoBERTa and pretrained through masked language modeling (MLM) on approximately 5~million  polymers augmented from PI1M \cite{ma_pi1m_2020}. It is fined-tuned on an augmented version of the DFT dataset shown in Table \ref{tab:table-dft} (with 7 out of 8 properties [no Eat]), which is about 2$\times$ $\sim$ 25$\times$ as large as the original size for each property. Data augmentation for both pretraining and fine-tuning is conducted by removing the canonical constraint in the SMILES representation and subsequently generating multiple noncanonical SMILES for each polymer structure. Polymer sequences are composed of polymer SMILES (of repeating units) and chemical descriptors and tokenized with a chemically aware tokenizer. The SMILES are split using a regular expression (regex), and each descriptor is treated as one token. 
    \item polyBERT \cite{kuenneth_polybert_2023} is based on DeBERTa and pretrained through MLM on approximately 80~million polymers generated from about 13,800 previously synthesized ones. 
    In fine-tuning, the pretrained transformer is frozen and used to provide polymer fingerprinting inputs to multitasking DNNs for property predictions. The DNNs are trained on a dataset with 29 properties from both experimental measurement and DFT (Table~\ref{tab:table-dft}). 
    These properties are split into six property categories---thermal, thermodynamic and  physical, electronic, optical and dielectric, mechanical, and permeability---and one multitasking DNN is trained for each category. Polymer sequences are represented by polymer SMILES (i.e., SMILES for repeating units with ``$\ast$''/``$[\ast]$'' for polymerization points) and tokenized with a regex-based tokenizer.
    \item SML-MT is our model based on BERT \cite{devlin2018bert}. It is pretrained through MLM on approximately 1~billion small molecules in Enamine REAL, similar to our previous work \cite{Blanchard2022}, and fine-tuned on the DFT dataset in Table~\ref{tab:table-dft}. In fine-tuning, the weights of the entire model (i.e., the BERT encoder followed by a linear NN layer with 8 neurons [one for each property]) are updated. Polymer sequences are represented by polymer SMILES with ``$\ast$'' for polymerization points and tokenized with a regex-based tokenizer.
\end{itemize}
Both the mean and standard deviation (STD) values of relative RMSE and R$^2$ for the test sets over the five folds in cross-validation are plotted in Figure \ref{fig:polar-R2RMSE}. 
For the 8 DFT properties, SML-MT outperforms TransPolymer and polyBERT for 5 (i.e., Egb, Eea, Eat, nc, and Ei), demonstrates comparable accuracy for 2 (i.e., Egc and eps), and has poor accuracy for 1 (i.e., Xc).
The poor prediction performance of all three models on Xc (i.e., crystallization tendency) is expected. Xc describes the tendency of a given polymer structure to form densified regions of ordered crystalline domains.
It is highly dependent on polymer structure descriptions across multiple length scales, and the current SMILES representation for only repeat units is inadequate.  

Our results show that a transformer pretrained on small molecules can achieve similar accuracy to the models trained on augmented polymer datasets. This suggests that a foundation model is a feasible approach to the data scarcity issue that plagues polymer informatics.

\section{Methods}

\subsection{Transformers for structure-property relationships}

We developed transformer-based models that take polymer sequence representations as input and then output property predictions. 
The polymer sequences are first split into tokens (a basic unit in LLMs) that can be processed by transformers via tokenization.
Subsequently, the tokenized sequences are fed into transformers with multihead attention layers that learn the correlation features embedded in polymer structures. 
Finally, the feature embeddings are mapped to properties via a linear NN layer. 
The following subsections describe the model and workflow details.

\paragraph{Polymer representation} 

The polymer-SMILES (p-SMILES) from PI1M \cite{ma_pi1m_2020} is used to represent polymers. It consists of SMILES for repeating units with ``$\ast$'' representing the polymerization points. 
More complicated representations that consider longer-chain structures with multiple repeating units are available, but their impact on prediction accuracy is unclear. For example, the single repeating unit has better performance than higher-degree polymerized representations for glass transition temperature predictions \cite{tao_benchmarking_2021}. In our work, we choose p-SMILES for its simplicity, sufficiency in representing the degree-1 polymers in PI1M, and similarity to SMILES representation of small molecules.%

\paragraph{Tokenization} 

Tokenization is the process of splitting SMILES (or p-SMILES) into smaller pieces (i.e., tokens), which are subsequently converted to numerical representations (via a vocabulary list) that are interpretable to transformers.  
We employ two splitting rules---regex (i.e., based on regex for individual atoms) and BERT (i.e., the default punctuation-based WordPiece tokenizer from Hugging Face \cite{wolf-etal-2020-transformers})---following our previous work \cite{Blanchard2022}.
The observations are similar for both rules. We will present the results with regex in the paper and provide the comparison with BERT in the Supplementary Materials (Figures \ref{fig:bar-RelativeRMSE-BERT}, \ref{fig:bar-R2-BERT}, and \ref{fig:token-regex-BERT}). 
For regex, the vocabulary list has 78 tokens consisting of 1--9, atoms, bonds, special tokens (i.e., [PAD], [UNK], [CLS], [SEP], and [MASK]), and ``$\ast$'' for polymerization points.

\paragraph{Model architecture} 

The transformer model used is BERT \cite{devlin2018bert} with MLM. We employed the standard \verb|BertForMaskedLM| implementation from Hugging Face with 12 hidden layers, 12 attention heads, and hidden size of 768. %
The pretraining is conducted with MLM (at a default 15\% masking), in which the model inputs are the preprocessed tokenized sequences (i.e., with [MASK] for masked tokens, prepended with [CLS], and appended with [SEP]), and the model is trained to predict the masked tokens.   
In fine-tuning for polymer property predictions, a dropout layer with a probability of 0.1 and a linear NN layer is attached to the [CLS] embedding output with the number of neurons equal to the number of properties. 

\paragraph{Transfer learning}

Transfer learning broadly refers to applying trained ML models to new applications that differ from the training data. In this work, we are interested in transferring the BERT model pretrained on small molecules to polymers. 
Training BERT requires large-scale %
datasets, which are well-documented for small molecules %
but not readily publicly available for polymers. %
Fundamentally, polymers and small molecules share basic chemical structure (which are transferable), but polymers have more complicated long-chain structures.  
It is thus unclear how much of the knowledge that the pretrained model learns from small molecules is transferable to polymers.  
In this work, we evaluate such transferability. To the best of our knowledge, this is the first example of transfer learning from from small molecules to polymers using a transformer architecture. 

\paragraph{Multitasking}

Multitasking, in this case predicting multiple properties simultaneously in a single model, is a popular strategy to leverage the potential correlations among properties for better data efficiency in deep learning.
It has been used to alleviate the data scarcity issue in polymers \cite{kuenneth2021polymer, gurnani2023polymer, kuenneth_polybert_2023}, including in polyBERT, for which 6 multitasking NNs were designed---with one for each property category. %
In our work, we train one multitasking model simultaneously for all properties.%

\subsection{Experiment settings}

The transformers are trained following the standard pretraining and fine-tuning procedure.

\subsubsection{Pretraining}

\begin{figure}[ht]
\centering
\includegraphics[width=0.5\linewidth]{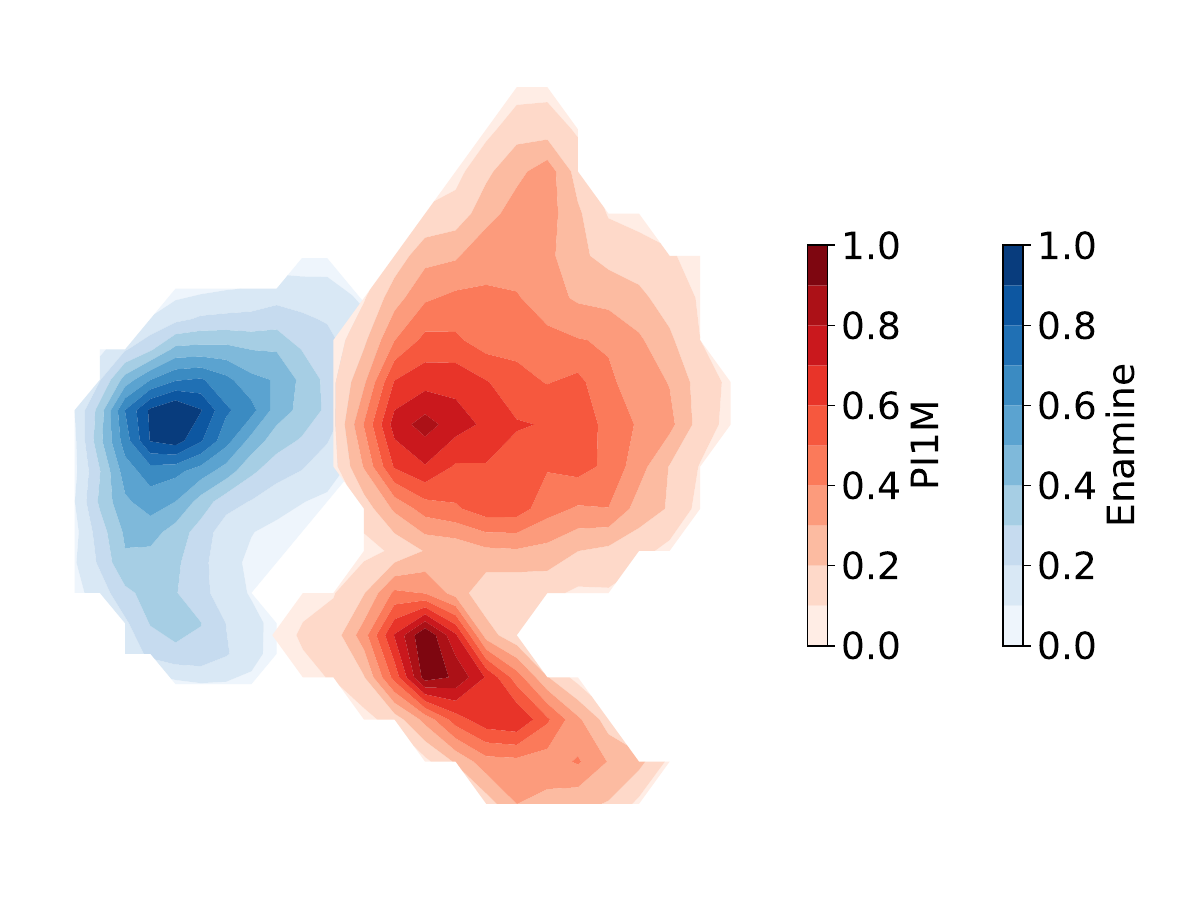}
\caption{Density plot of extended-connectivity fingerprints projected on the first two principal components for structures in PI1M and Enamine REAL. It shows that the two datasets contain very different structures, thereby presenting potential challenges in the transfer learning task. %
}
\label{fig:umap}
\end{figure}

\paragraph{Datasets} To evaluate the transferability of transformers from small molecules to polymers, we consider two datasets in pretraining: Enamine REAL for small molecules and PI1M for polymers.
Enamine REAL is a popular compound library \cite{shivanyuk2007enamine} for deep learning studies \cite{Blanchard2022,acharya2020supercomputer,blanchard_SARS-CoV-2-2022,blanchard2023adaptive}.
The dataset used here is taken from our previous work \cite{Blanchard2022} and consists of approximately 1~billion small molecules.
PI1M consists of approximately 1~million degree-1 polymers generated from PolyInfo, which has also been used in multiple deep learning studies \cite{tao_machine_2021, yang2022machine, xu2023transpolymer}, including for TransPolymer.
To compare the molecular structures in two datasets, we calculate the extended-connectivity fingerprints (ECFPs) with 1,024 bits from SMILES using RDKit.
Figure~\ref{fig:umap} shows the data density plot of the ECFPs projected on the first two principal components (PCs) in a principal component analysis for 50,000 randomly selected samples from the two datasets. 
The density plot shows that the two datasets have very different ECFP distributions and presents potential challenges in the transfer learning task.

\paragraph{Training on Summit} 
The two models pretrained on Enamine REAL and PI1M are referred to as SML (small) and PLM (polymer), respectively, as shown in Table~\ref{tab:pretrainedmodel}. The model without pretraining (i.e., INIT [randomly initialized]) is included as a reference. We use the tokenizer and BERT with MLM implementations from Hugging Face \cite{wolf-etal-2020-transformers} along with the DeepSpeed optimization library \cite{Rajbhandari_deepspeed_2020}. 
In training, the learning rate was set to $10^{-3}$. The WarmupDecayLR scheduler was set with a warmup ratio of 0.1 and the LAMB optimizer \cite{you2019large},  and the batch size was set to 64 per GPU. 
The models were trained for 20 epochs with gradient accumulation every 3 steps with 16-bit mixed-precision training. 
All training runs were performed on Summit (with 6 NVIDIA V100 GPUs for each node) at the Oak Ridge Leadership Computing Facility.
The SML training used 334 nodes and took 6.1 hours on about 1~billion SMILES strings.
PLM was trained using 1 node for 2.6 hours on about 1~million SMILES strings.

\begin{table}[h!]
  \begin{center}
    \caption{Pretrained models}
    \label{tab:pretrainedmodel}
    \begin{tabular}{l|c|c|c} 
    \textbf{Dataset}& Small molecule (Enamine REAL) &PI1M & No pretraining \\
   \hline
    \textbf{Model} & SML & PLM & INIT \\
    \end{tabular}
  \end{center}
\end{table}

\begin{table}[h!]
  \begin{center}
    \caption{Fine-tuned models}
    \label{tab:finetunedmodel}
    \begin{tabular}{l|c|c|c} 
      \textbf{Model}&SML & PLM & INIT  \\
      \hline
      Single-tasking (ST)&SML-ST&PLM-ST&INIT-ST\\
      Multitasking (MT)&SML-MT&PLM-MT&INIT-MT\\
    \end{tabular}
  \end{center}
\end{table}

\subsubsection{Fine-tuning}
In fine-tuning, the three pretrained models, SML, PLM, and INIT (with one dropout layer and one linear NN layer attached in the end for property outputs), are trained on the polymer DFT dataset in Table~\ref{tab:table-dft}. In total, six types of models are trained depending on the single-tasking and multitasking context summarized in Table \ref{tab:finetunedmodel}; %
or equivalently, $27=3\times(8+1)$ models are trained, where 3 is for the 3 pretrained models, %
8 is for the 8 single-tasking models with one for each property, and 1 is for the multitasking model for all the DFT properties. 

For each regression task, the dataset in Table \ref{tab:table-dft} was split into 5 folds for cross-validation; therefore, 5 models were trained. By default, the AdamW optimizer was used with a learning rate of $10^{-5}$ and zero weight decay, the batch size was set to 16, and the training runs were performed on one GPU for 600 epochs with L1 loss function.

\subsubsection{Model evaluation}

The models (Table \ref{tab:finetunedmodel}) learn the mappings from sequence space to feature embedding space and subsequently to property space.
In pretraining, the model identifies a mapping from structure space to embedding space that is close to the mapping (of interest for downstream tasks) by learning generic knowledge from diverse, large-scale structure data. In fine-tuning, the model adjusts the mapping and simultaneously learns the subsequent mapping to property space with a low-cost iterative training on structure-property data. %
The final model accuracy is jointly determined by knowledge learned in both stages.
A good pretrained model can lead to a faster convergence to the final solutions in fine-tuning. 
To understand the feasibility of transferring transformers pretrained on small molecules to polymers, we will evaluate not only the final model accuracy but also the quality of pretraining. 

For the prediction accuracy, we will calculate both the R$^2$ and relative RMSE with STD from the 5-fold cross-validation and compare them with the state-of-the-art values from TransPolymer \cite{xu2023transpolymer} and polyBERT\cite{kuenneth_polybert_2023}.  
For the pretraining quality, we will evaluate the pretrained models in (1) the MLM tasks for prediction accuracy of the masked tokens in test sets from Enamine REAL and PI1M and (2) the fine-tuning tasks (i.e., prediction of DFT properties) for the convergent behaviors by varying the training set size from zero-short learning (i.e., with no fine-tuning) to full-scale training with all training samples.

\section{Results and discussions}

\subsection{Prediction accuracy of fine-tuned models on DFT properties}

We evaluate the predictive accuracy of the models for polymer properties pretrained on Enamine REAL (SML) and PI1M (PLM) and fine-tuned for individual property (single-tasking [ST]) and all properties simultaneously (multitasking [MT]). These models are compared with the two state-of-the-art models: TransPolymer \cite{xu2023transpolymer} and polyBERT \cite{kuenneth_polybert_2023}. 
Figure \ref{fig:polar-R2RMSE} shows the results of SML-MT, which achieves accuracy comparable with both TransPolymer and polyBERT. In the following, we present the evaluation results of all models in Table \ref{tab:finetunedmodel}.
Both the relative RMSE and the R$^2$ from a 5-fold cross-validation are calculated. We present relative RMSE only, and the corresponding R$^2$ results are available in the Supplementary Material.

\begin{figure}[h]
\centering
\includegraphics[width=0.9\linewidth]{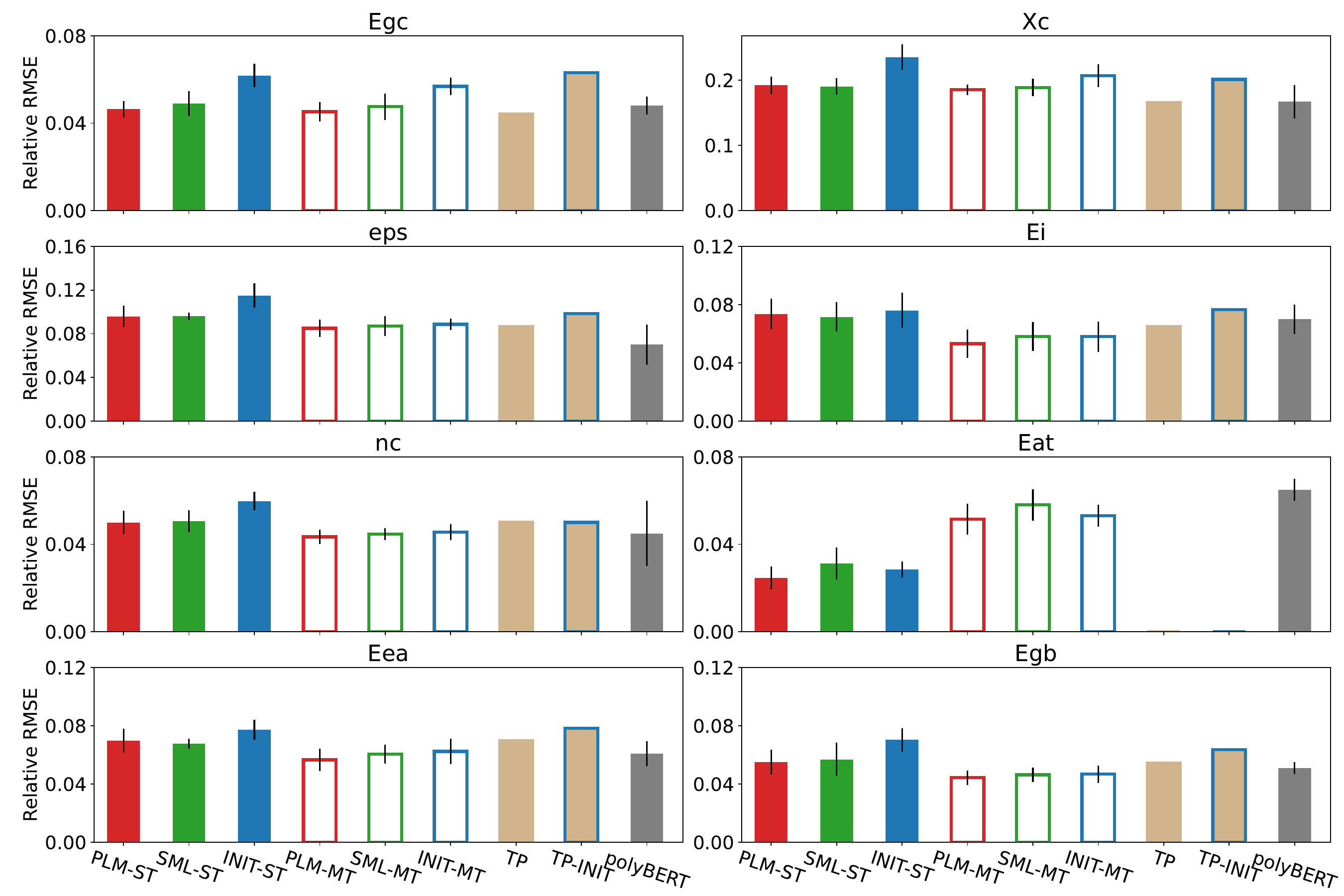}
\caption{Relative RMSE prediction error of DFT properties. Error bars show the STD from the 5-fold cross-validation. Results of TransPolymer \cite{xu2023transpolymer} (with no STD reported and hence no error bars plotted), where TP and TP-INIT represent the models with and without pretraining, respectively, and polyBERT \cite{kuenneth_polybert_2023} are also shown for reference. Pretraining increases model prediction accuracy for downstream tasks.
The best pretrained models achieve accuracy comparable to or better than TransPolymer and polyBERT. Multitasking (MT, empty bars) shows an obvious accuracy improvement for eps, Ei, nc, Eea, and Egb; comparable accuracy for Egc and Xc; and degraded accuracy for Eat versus single-tasking (ST, solid bars).} 
\label{fig:bar-RMSE}
\end{figure}

Figure \ref{fig:bar-RMSE} shows the bar plot of the mean (over 5 folds) relative RMSE values for the test sets along with the TransPolymer (TP), TransPolymer without pretraining (TP-INIT), and polyBERT results reproduced from the literature \cite{xu2023transpolymer, kuenneth_polybert_2023}. The error bars represent the STD over 5 runs. 
The models without pretraining (i.e., INIT and bars with blue edge color) in general show higher predictive errors than the models pretrained on Enamine REAL and PI1M. This is consistent in both ST (solid bars) and MT (empty bars).
The best models from ST and MT obtain accuracy comparable to or better than TransPolymer and polyBERT. 
The models pretrained on Enamine REAL (i.e., SML and green bars) have similar predictive errors to the models pretrained on PI1M (i.e., PLM and red bars) in both ST and MT. This, together with Figure \ref{fig:polar-R2RMSE}, supports the idea that BERT \cite{devlin2018bert} trained on small molecules works as a foundation model for polymers.
The numerical values for the bar plots are available in Table~\ref{tab:finetuning-RelativeRMSE} in the Supplementary Material, where the results of the training sets are also presented; the two best models for each property are highlighted in bold.

For the comparison between ST (solid bars) and MT (empty bars), the detailed performance results are property-dependent, although MT generally shows better overall performance than ST.  
In particular, MT achieves an obvious accuracy improvement for eps, Ei, nc, Eea, and Egb; comparable accuracy for Egc and Xc; and degraded accuracy for Eat versus ST.
This is consistent with the observations reported in the polymer property prediction work \cite{kuenneth2021polymer} with chemistry-informed Polymer Genome fingerprint embeddings (that considers hierarchical length scales), where ST performed slightly better than MT on thermodynamic and physical properties (i.e., Eat and Xc). %
The different performance of ST and MT across properties is a combination of available training data and the potential physical correlations between them. 
In general, the different tasks in MT interact in a collaborative and competing manner---highly correlated (uncorrelated) ones assist (compete with) each other.
For the polymer training data in Table \ref{tab:table-dft}, strong overlapping across properties exists. That is, given a polymer SMILES, there can be multiple properties. Table~\ref{tab:num_samples} shows such overlapping. For the 390 polymers in Eat (a thermodynamic and physical property showing decreased accuracy from ST to MT), more than 91\% share the same structures with 6 properties from the other two distinct categories. The learning task is expected to be more difficult in MT, and, consequently, ST is preferred for better accuracy.

\begin{table}[h]
  \begin{center}
    \caption{Number of overlapping samples across DFT properties in Table \ref{tab:table-dft}}
    \label{tab:num_samples}
    \resizebox{0.8\textwidth}{!}{
    \begin{tabular}{l|ccccccccc} 
\textbf{Property}&Eat &Xc &Egc &Egb &Eea &Ei&nc&eps \\\hline
Eat (thermodynamic and physical)&390 &   &   &   &   &   &   &  \\
Xc (thermodynamic and physical)&15 & 432 &   &   &   &   &   &  \\
Egc (electronic)&365 & 191 & 3380 &   &   &   &   &  \\
Egb (electronic)&358 & 31 & 543 & 561 &   &   &   &  \\
Eea (electronic)&366 & 14 & 351 & 343 & 368 &   &   &  \\
Ei (electronic)&368 & 15 & 353 & 345 & 368 & 370 &   &  \\
nc (optical and dielectric)&380 & 15 & 360 & 353 & 361 & 363 & 382 &  \\
eps (optical and dielectric)&380 & 15 & 360 & 353 & 361 & 363 & 382 & 382\\
 \end{tabular}}
  \end{center}
\end{table}

\subsection{Quality of pretraining}

\subsubsection{Prediction accuracy of pretrained models on masked tokens in MLM}

In pretraining, the PLM and SML models are trained with the standard MLM strategy with a 15\% masking rate on PI1M (PLM) and Enamine REAL (SML). Two corresponding test sets that consist of 9,999 SMILES strings unseen during training are used to evaluate model accuracy in predicting masked tokens.

\begin{figure}[h]
\centering
\includegraphics[width=0.45\linewidth]{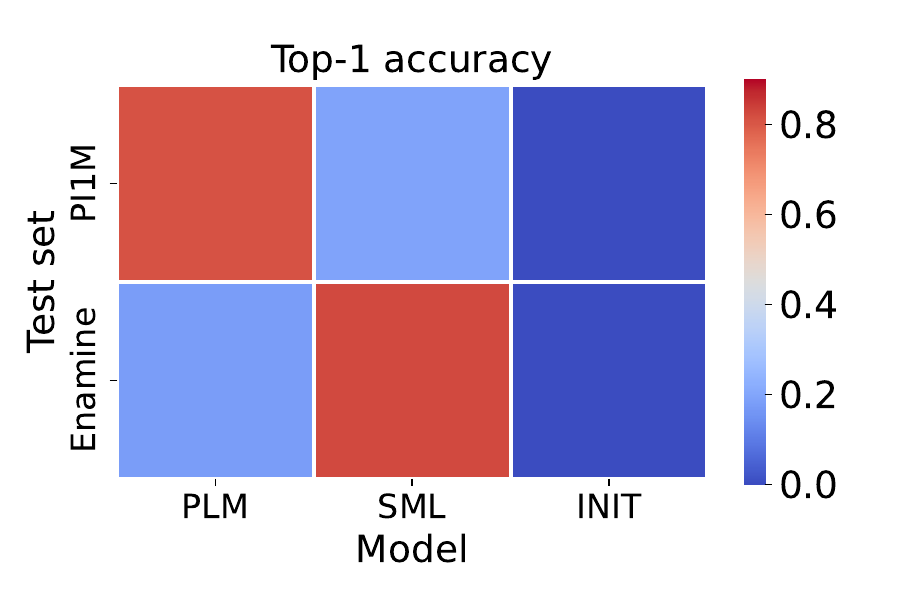}
\caption{Top-1 accuracy of token predictions in MLM of models PLM, SML, and INIT for PI1M and Enamine test sets with 9,999 SMILES. }
\label{fig:heatmap-pretrained}
\end{figure}

Figure \ref{fig:heatmap-pretrained} shows the top-1 accuracy of PLM and SML together with INIT (i.e., without pretraining) on the two test sets. 
The pretrained PLM and SML models perform well in the in-distribution test sets PI1M and Enamine, respectively, with top-1 test accuracy higher than 81\%.
Without pretraining, the INIT model fails to predict any SMILES correctly.
In contrast, the cross-testing accuracy (i.e., PLM tested on Enamine and SML tested on PI1M) is about 20\%, which indicates the pretrained models PLM and SML learned transferable structure information between polymers and small molecules.

\begin{figure}[ht]
\centering
\includegraphics[width=0.75\linewidth]{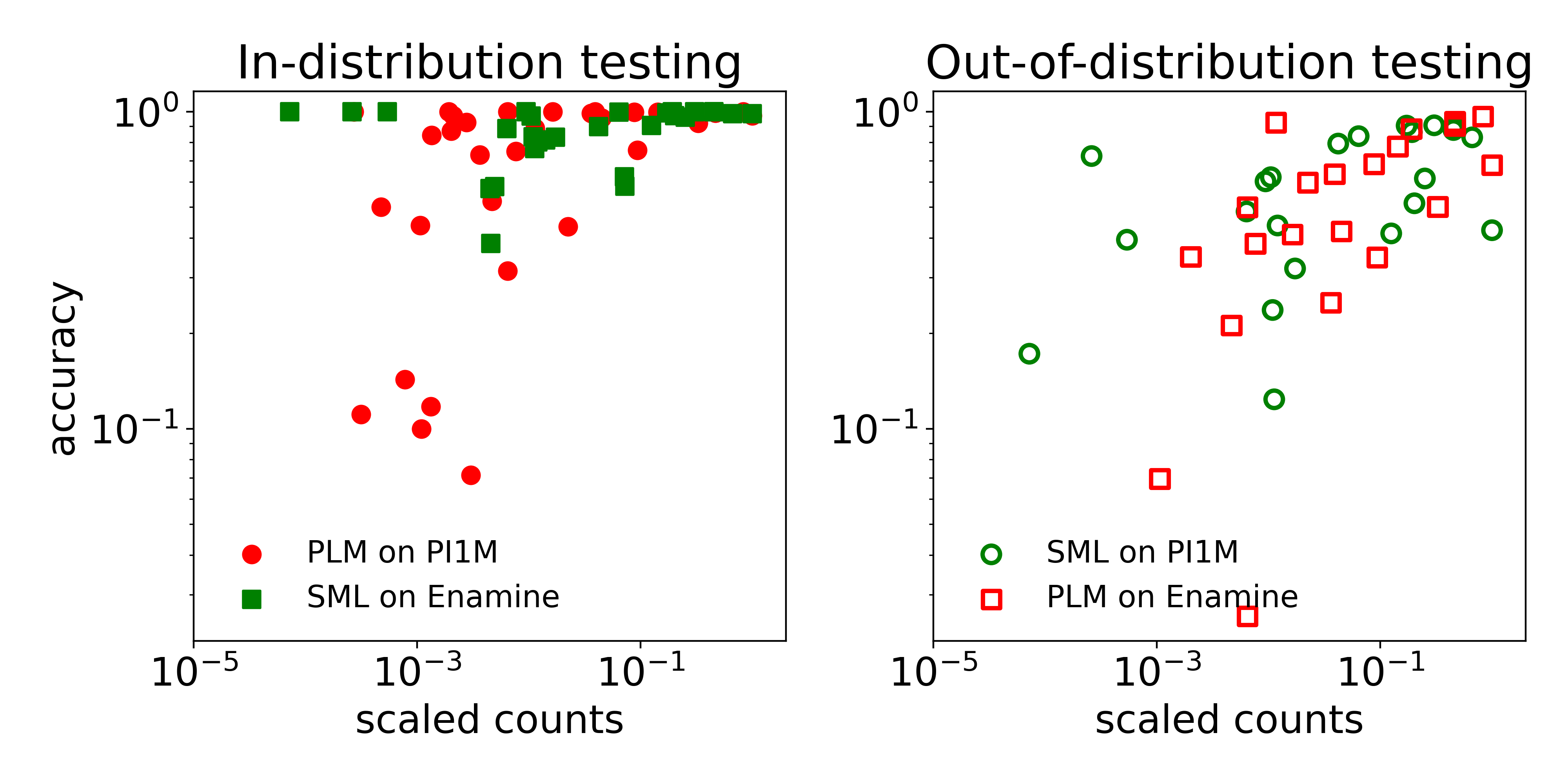}
\caption{Prediction accuracy of masked tokens in the test sets against the scaled frequency in the training sets. The tokens with higher frequencies in the training sets are generally better captured.}
\label{fig:tokens-accuracy-counts}
\end{figure}
We examine the prediction accuracy of individual tokens in MLM.
Figure~\ref{fig:tokens-accuracy-counts} shows the prediction accuracy of masked tokens from the test sets versus the corresponding scaled occurrence frequency in the training sets. In general, the ones with higher frequencies in training sets are better captured for the in-distribution testing (i.e., models are trained and tested on similar structures). %
Even for the (out-of-distribution) cross-testing, in which the models are trained and tested on very different structures (Figure~\ref{fig:umap}), a more scattered but still evident correlation can be seen. 
For \textit{C}, which is the most frequent token (with the scaled counts of 1.0) in the training sets, the models achieve decent accuracy of around $0.4\sim0.7$ in the out-of-distribution testing.
This again indicates generic transferable knowledge learned by the models and hence supports the foundation model approach.
In the Supplementary Materials, we show the top-20 most and least frequent tokens (Figure \ref{fig:tokens-counts}), the top-20 most and least accurately predicted tokens (Figure \ref{fig:tokens-accuracy}), and a few examples of successfully and falsely predicted SMILES strings (Figure \ref{fig:SMILES-crossing}) from the out-of-distribution testing.

\subsubsection{Learning efficiency of pretrained models in fine-tuning}

The prediction accuracy shown in Figure \ref{fig:bar-RMSE} is determined by both pretraining and fine-tuning---pretraining aims to set up a good initial guess of model weights by learning generic knowledge from a large-scale structure dataset, and fine-tuning further refines them for specific properties with structure-property data. However, only looking at the final accuracy to evaluate the quality of pretraining can be misleading; recent examples have shown that higher pretraining accuracy can saturate or decrease fine-tuning accuracy \cite{abnar2021exploring}. Different pretrained models may show equally good prediction accuracy when the structure-property data is adequate in both quantity and quality; on the other hand, all models would show equally poor accuracy when there is no structure-property relationship.
What sets a pretrained model apart is learning efficiency: the amount of structure-property data (which is expensive to collect) required to achieve the same prediction accuracy. 

\begin{figure}[!ht]
\centering
\includegraphics[width=0.9\linewidth]{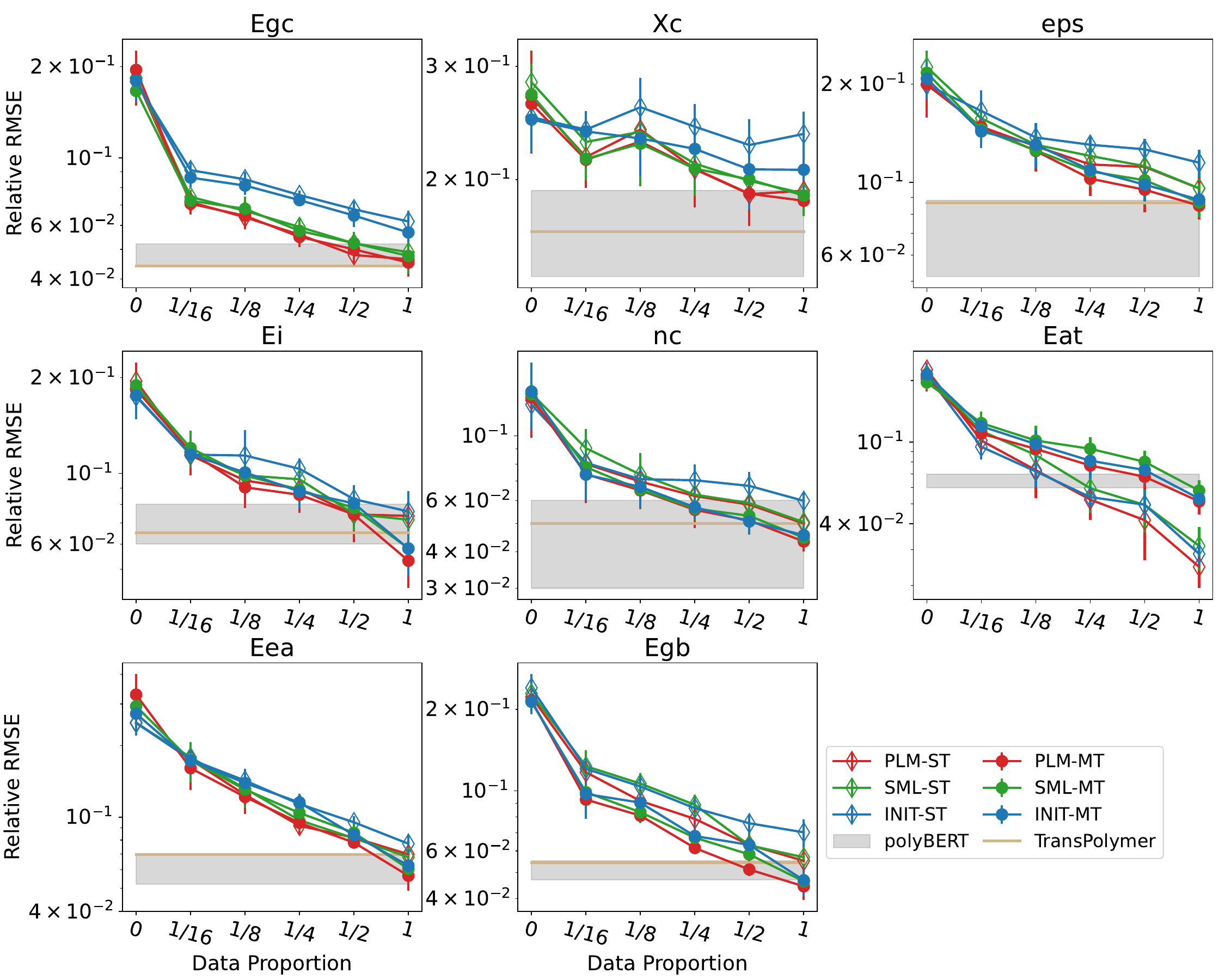}
\caption{Relative RMSE prediction error of DFT properties at varying training set sizes ([0, 1/16, 1/8, 1/4, 1/2, 1] of the full-scale size) in fine-tuning. Results of TransPolymer \cite{xu2023transpolymer} (with brown lines for the mean relative RMSE) and polyBERT \cite{kuenneth_polybert_2023} (with gray shadow areas for the $\text{mean}\pm1\ \text{STD}$ of relative RMSE from the 5-fold cross-validation) are also shown for reference. Figure~\ref{fig:bar-RMSE} corresponds to results for a data proportion of 1.0. The two pretrained models PLM (in red) and SML (in green) show better learning efficiency than the INIT model without pretraining (in blue) and hence require less expensive structure-property data for the same accuracy. Multitasking (MT, solid circles) improves prediction accuracy for most variables versus single-tasking (ST, empty diamonds), especially for models without pretraining (INIT in blue).
}
\label{fig:profile-RMSE}
\end{figure}

Figure \ref{fig:profile-RMSE} shows the learning efficiency of three pretrained models (SML, PLM, and INIT) in both ST and MT. In total, six sets of fine-tuning runs were conducted with the training set sizes at [0, 1/16, 1/8, 1/4, 1/2, 1] of the full-scale size. The STD from the 5-fold cross-validation is shown as error bars.      
Results of TransPolymer \cite{xu2023transpolymer} and polyBERT \cite{kuenneth_polybert_2023} are also shown for reference, where the brown lines show the mean relative RMSE for TransPolymer, and the gray shadow areas show the $\text{mean}\pm1\ \text{STD}$ of relative RMSE from the 5-fold cross-validation for polyBERT. 
For zero-shot learning at a data proportion of 0.0, all the models perform equally poorly because they are agnostic of property outputs with the weights of final NN layer initialized randomly following a normal distribution.
With the training size increasing (i.e., the models being exposed to more structure-property examples), some pretrained models start to stand out by showing higher learning efficiencies with lower relative RMSE errors, such as the SML and PLM shown in green and red (in contrast to INIT shown in blue) for Egc. Moreover, although the relative RMSEs of the SML models are marginally higher than for PLM, they are still comparable. This further supports the transfer learning idea from small molecules to polymers, especially when considering the substantial cost saved in data augmentation and pretraining.

Similar trends hold for the other properties, although they may not be as evident as for Egc. For example, in Xc, the difference between the two pretrained SML and PLM models and INIT increases from 1/8 to 1 but is associated with large overlapping error bars.   
The different behaviors across properties could be caused by data size. Notably, Egc has several times more samples than the other properties listed in Table \ref{tab:table-dft}. 

Regarding the learning efficiency for ST (empty symbols) and MT (solid symbols), similar observations can be made of the data plotted in Figure~\ref{fig:bar-RMSE}: compared to ST, MT shows an obvious improvement for eps, Ei, nc, Eea, and Egb; is comparable for Egc and Xc; and degrades for Eat. The different performance across properties is caused by the training data available for each property and the correlations with others in the collaborative-competitive MT, as detailed in the previous section (Figure \ref{fig:bar-RMSE}).   
Moreover, the improvement of MT is more prominent in INIT (with a larger difference observed between solid blue circles and empty blue diamonds). For some properties (e.g., eps and nc), INIT-MT even achieves accuracy comparable to the pretrained models. This suggests that both MT and pretraining are promising techniques to improve prediction accuracy.

\section{Conclusion and future work}

To summarize, we proposed a foundation model approach to address the data scarcity issue in polymer informatics. 
In particular, we pretrained a transformer-based model on the Enamine REAL large-scale small molecule dataset, and showed that the model can achieve accuracy comparable with the TransPolymer and polyBERT models without the need for expensive generation and pretraining on large-scale augmented polymer datasets. 
This further supports the potential of using a model pretrained on a large-scale, accessible, and general molecular dataset as a foundation model for multiple applications across the molecular design landscape.

We compared the models pretrained on Enamine REAL and PI1M with models without pretraining in ST and MT contexts for prediction accuracy on polymer DFT properties and pretraining quality.
The models pretrained on the two datasets have similar predictive behaviors: (1) they showed comparable prediction accuracy on polymer properties in ST and MT (consistently higher than models without pretraining); (2) they both achieved about 20\% prediction accuracy on masked tokens in pretraining with MLM in cross-testing (in contrast to the 0\% accuracy without pretraining ); and (3) they exhibited similar learning efficiency in the convergence test on the training set size in fine-tuning.
All these empirical observations support the feasibility of transfer learning from a model pretrained on small molecules to polymers. 
Moreover, MT enhances the prediction accuracy in general, even though the detailed performance is property-dependent, as a result of its collaborative-competitive nature. 

Although our foundation model approach achieved promising results by accurately predicting local DFT properties that explicitly depend on bond connectivity of the repeat units, we do not expect this to be directly applicable to all polymeric properties, especially those that invoke longer-range dependencies. %
This is reflected in the models' poor prediction performance for Xc, which describes the tendency of a given polymer structure to form densified regions of ordered crystalline domains. %
Excluding that Xc is highly dependent on the thermomechanical history of the polymer, it is highly mediated by an ensemble of properties that require descriptions at multiple length scales (e.g., polymer chain flexibility, stereoregularity, and the intensity of repeat unit intermolecular interactions). %
These properties are not adequately encoded within a SMILES string but categorically mediate the dynamics of the macromolecules preceding solidification. 
This issue is common to all language models (including ours, TransPolymer, and polyBERT) that employ SMILES representations for polymers. %
To improve the prediction of these properties (e.g., Xc, glass transition temperature), a future research direction would be to incorporate additional chemistry-informed layers of increasingly longer-range descriptors (e.g., degree of polymerization, connectivity) that pertain to topology in polymer representations.

\section{Data and software availability}

All datasets used to train and test the models are open source. Enamine REAL \cite{shivanyuk2007enamine} can be found at https://enamine.net/compound-collections/real-compounds/real-database/. PI1M \cite{ma_pi1m_2020} was downloaded from https://github.com/RUIMINMA1996/PI1M/, and the DFT property dataset \cite{kuenneth2021polymer} was downloaded from https://khazana.gatech.edu/.

As for the codes, we used the \verb|BertForMaskedLM| Hugging Face implementation of the BERT model with the MLM.
The pretrained and fine-tuned models are available at Hugging Face https://huggingface.co/mossaic-candle/.
\begin{acknowledgement}


The research described here was supported by the Exascale Computing Project (17-SC-20-SC), a collaborative effort of the US Department of Energy Office of Science and the National Nuclear Security Administration. This research used resources of the Oak Ridge Leadership Computing Facility, which is a DOE Office of Science User Facility supported under Contract DE-AC0500OR22725. LK and AKN acknowledge support from US Department of Energy, Office of Science, Basic Energy Sciences under contract ERKCK60.

This manuscript has been authored by UT-Battelle LLC under contract DE-AC05-00OR22725 with the US Department of Energy (DOE). The US government retains and the publisher, by accepting the article for publication, acknowledges that the US government retains a nonexclusive, paid-up, irrevocable, worldwide license to publish or reproduce the published form of this manuscript, or allow others to do so, for US government purposes. DOE will provide public access to these results of federally sponsored research in accordance with the DOE Public Access Plan (https://www.energy.gov/doe-public-access-plan).

\end{acknowledgement}

\begin{suppinfo}


The following files are available free of charge.

\begin{itemize}
  \item SM\_1.pdf: More prediction results of fine-tuned models in polymer properties, including bar plots of R$^2$ in Figure \ref{fig:bar-R2}, tables of relative RMSE and R$^2$ in Tables \ref{tab:finetuning-RelativeRMSE} and \ref{tab:finetuning-R2}, and parity plots of the training and test sets from the 5-fold cross-validation in Figure \ref{fig:parityplot}. 
  \item SM\_2.pdf: More results of pretrained models in MLM, including the top-20 most and least frequent tokens in Figure \ref{fig:tokens-counts}, top-20 most and least accurately predicted tokens in Figure \ref{fig:tokens-accuracy}, and a few examples of successfully and falsely predicted SMILES strings in Figure \ref{fig:SMILES-crossing} from the out-of-distribution testing.
  \item SM\_3.pdf: Prediction results of models with BERT tokenization method, including bar plots of relative RMSE in Figure \ref{fig:bar-RelativeRMSE-BERT}, R$^2$ in Figure \ref{fig:bar-R2-BERT}, and prediction performance comparison of models with BERT and regex tokenizers in Figure \ref{fig:token-regex-BERT}. 
\end{itemize}

\end{suppinfo}


\begin{mcitethebibliography}{47}
\providecommand*\natexlab[1]{#1}
\providecommand*\mciteSetBstSublistMode[1]{}
\providecommand*\mciteSetBstMaxWidthForm[2]{}
\providecommand*\mciteBstWouldAddEndPuncttrue
  {\def\EndOfBibitem{\unskip.}}
\providecommand*\mciteBstWouldAddEndPunctfalse
  {\let\EndOfBibitem\relax}
\providecommand*\mciteSetBstMidEndSepPunct[3]{}
\providecommand*\mciteSetBstSublistLabelBeginEnd[3]{}
\providecommand*\EndOfBibitem{}
\mciteSetBstSublistMode{f}
\mciteSetBstMaxWidthForm{subitem}{(\alph{mcitesubitemcount})}
\mciteSetBstSublistLabelBeginEnd
  {\mcitemaxwidthsubitemform\space}
  {\relax}
  {\relax}

\bibitem[Audus and de~Pablo(2017)Audus, and de~Pablo]{audus2017polymer}
Audus,~D.~J.; de~Pablo,~J.~J. Polymer informatics: Opportunities and
  challenges. \emph{ACS macro letters} \textbf{2017}, \emph{6},
  1078--1082\relax
\mciteBstWouldAddEndPuncttrue
\mciteSetBstMidEndSepPunct{\mcitedefaultmidpunct}
{\mcitedefaultendpunct}{\mcitedefaultseppunct}\relax
\EndOfBibitem
\bibitem[Cencer \latin{et~al.}(2022)Cencer, Moore, and
  Assary]{cencer_machine_2022}
Cencer,~M.~M.; Moore,~J.~S.; Assary,~R.~S. Machine learning for polymeric
  materials: an introduction. \emph{Polymer International} \textbf{2022},
  \emph{71}, 537--542, Publisher: John Wiley and Sons Ltd\relax
\mciteBstWouldAddEndPuncttrue
\mciteSetBstMidEndSepPunct{\mcitedefaultmidpunct}
{\mcitedefaultendpunct}{\mcitedefaultseppunct}\relax
\EndOfBibitem
\bibitem[Chen \latin{et~al.}(2021)Chen, Pilania, Batra, Huan, Kim, Kuenneth,
  and Ramprasad]{chen_polymer_2021}
Chen,~L.; Pilania,~G.; Batra,~R.; Huan,~T.~D.; Kim,~C.; Kuenneth,~C.;
  Ramprasad,~R. Polymer informatics: {Current} status and critical next steps.
  \emph{Materials Science and Engineering: R: Reports} \textbf{2021},
  \emph{144}, 100595\relax
\mciteBstWouldAddEndPuncttrue
\mciteSetBstMidEndSepPunct{\mcitedefaultmidpunct}
{\mcitedefaultendpunct}{\mcitedefaultseppunct}\relax
\EndOfBibitem
\bibitem[Tao \latin{et~al.}(2021)Tao, Varshney, and Li]{tao_benchmarking_2021}
Tao,~L.; Varshney,~V.; Li,~Y. Benchmarking {Machine} {Learning} {Models} for
  {Polymer} {Informatics}: {An} {Example} of {Glass} {Transition}
  {Temperature}. \emph{Cite This: J. Chem. Inf. Model} \textbf{2021},
  \emph{61}, 5413\relax
\mciteBstWouldAddEndPuncttrue
\mciteSetBstMidEndSepPunct{\mcitedefaultmidpunct}
{\mcitedefaultendpunct}{\mcitedefaultseppunct}\relax
\EndOfBibitem
\bibitem[Jablonka \latin{et~al.}(2021)Jablonka, Jothiappan, Wang, Smit, and
  Yoo]{jablonka_bias_2021}
Jablonka,~K.~M.; Jothiappan,~G.~M.; Wang,~S.; Smit,~B.; Yoo,~B. Bias free
  multiobjective active learning for materials design and discovery.
  \emph{Nature Communications} \textbf{2021}, \emph{12}, 1--10, Publisher:
  Springer US\relax
\mciteBstWouldAddEndPuncttrue
\mciteSetBstMidEndSepPunct{\mcitedefaultmidpunct}
{\mcitedefaultendpunct}{\mcitedefaultseppunct}\relax
\EndOfBibitem
\bibitem[Kuenneth \latin{et~al.}(2022)Kuenneth, Lalonde, Marrone, Iverson,
  Ramprasad, and Pilania]{kuenneth2022bioplastic}
Kuenneth,~C.; Lalonde,~J.; Marrone,~B.~L.; Iverson,~C.~N.; Ramprasad,~R.;
  Pilania,~G. Bioplastic design using multitask deep neural networks.
  \emph{Communications Materials} \textbf{2022}, \emph{3}, 96\relax
\mciteBstWouldAddEndPuncttrue
\mciteSetBstMidEndSepPunct{\mcitedefaultmidpunct}
{\mcitedefaultendpunct}{\mcitedefaultseppunct}\relax
\EndOfBibitem
\bibitem[Nagasawa \latin{et~al.}(2018)Nagasawa, Al-Naamani, and
  Saeki]{nagasawa_computer-aided_2018}
Nagasawa,~S.; Al-Naamani,~E.; Saeki,~A. Computer-{Aided} {Screening} of
  {Conjugated} {Polymers} for {Organic} {Solar} {Cell}: {Classification} by
  {Random} {Forest}. \emph{The Journal of Physical Chemistry Letters}
  \textbf{2018}, \emph{9}, 2639--2646, Publisher: American Chemical
  Society\relax
\mciteBstWouldAddEndPuncttrue
\mciteSetBstMidEndSepPunct{\mcitedefaultmidpunct}
{\mcitedefaultendpunct}{\mcitedefaultseppunct}\relax
\EndOfBibitem
\bibitem[Kuenneth \latin{et~al.}(2021)Kuenneth, Schertzer, and
  Ramprasad]{kuenneth_copolymer_2021}
Kuenneth,~C.; Schertzer,~W.; Ramprasad,~R. Copolymer {Informatics} with
  {Multitask} {Deep} {Neural} {Networks}. \emph{Macromolecules} \textbf{2021},
  \emph{54}, 5957--5961, arXiv: 2103.14174\relax
\mciteBstWouldAddEndPuncttrue
\mciteSetBstMidEndSepPunct{\mcitedefaultmidpunct}
{\mcitedefaultendpunct}{\mcitedefaultseppunct}\relax
\EndOfBibitem
\bibitem[Sattari \latin{et~al.}(2021)Sattari, Xie, and
  Lin]{sattari_data-driven_2021}
Sattari,~K.; Xie,~Y.; Lin,~J. Data-driven algorithms for inverse design of
  polymers. \emph{Soft Matter} \textbf{2021}, \emph{17}, 7607--7622, Publisher:
  Royal Society of Chemistry\relax
\mciteBstWouldAddEndPuncttrue
\mciteSetBstMidEndSepPunct{\mcitedefaultmidpunct}
{\mcitedefaultendpunct}{\mcitedefaultseppunct}\relax
\EndOfBibitem
\bibitem[Ma \latin{et~al.}(2019)Ma, Liu, Zhang, Liu, and
  Luo]{ma_evaluating_2019}
Ma,~R.; Liu,~Z.; Zhang,~Q.; Liu,~Z.; Luo,~T. Evaluating {Polymer}
  {Representations} via {Quantifying} {Structure}–{Property} {Relationships}.
  \emph{Journal of Chemical Information and Modeling} \textbf{2019}, \emph{59},
  3110--3119, Publisher: American Chemical Society\relax
\mciteBstWouldAddEndPuncttrue
\mciteSetBstMidEndSepPunct{\mcitedefaultmidpunct}
{\mcitedefaultendpunct}{\mcitedefaultseppunct}\relax
\EndOfBibitem
\bibitem[Jha \latin{et~al.}(2019)Jha, Chandrasekaran, Kim, and
  Ramprasad]{jha_impact_2019}
Jha,~A.; Chandrasekaran,~A.; Kim,~C.; Ramprasad,~R. Impact of dataset
  uncertainties on machine learning model predictions: the example of polymer
  glass transition temperatures. \emph{Modelling and Simulation in Materials
  Science and Engineering} \textbf{2019}, \emph{27}, 024002, Publisher: IOP
  Publishing\relax
\mciteBstWouldAddEndPuncttrue
\mciteSetBstMidEndSepPunct{\mcitedefaultmidpunct}
{\mcitedefaultendpunct}{\mcitedefaultseppunct}\relax
\EndOfBibitem
\bibitem[Tao \latin{et~al.}(2021)Tao, Chen, and Li]{tao_machine_2021}
Tao,~L.; Chen,~G.; Li,~Y. Machine learning discovery of high-temperature
  polymers. \emph{Patterns} \textbf{2021}, \emph{2}, 100225, Publisher:
  Elsevier Inc.\relax
\mciteBstWouldAddEndPunctfalse
\mciteSetBstMidEndSepPunct{\mcitedefaultmidpunct}
{}{\mcitedefaultseppunct}\relax
\EndOfBibitem
\bibitem[Doan~Tran \latin{et~al.}(2020)Doan~Tran, Kim, Chen, Chandrasekaran,
  Batra, Venkatram, Kamal, Lightstone, Gurnani, Shetty, Ramprasad, Laws,
  Shelton, and Ramprasad~AFFILIATIONS]{doan_tran_machine-learning_2020}
Doan~Tran,~H.; Kim,~C.; Chen,~L.; Chandrasekaran,~A.; Batra,~R.; Venkatram,~S.;
  Kamal,~D.; Lightstone,~J.~P.; Gurnani,~R.; Shetty,~P.; Ramprasad,~M.;
  Laws,~J.; Shelton,~M.; Ramprasad~AFFILIATIONS,~R. Machine-learning
  predictions of polymer properties with {Polymer} {Genome}. \emph{J. Appl.
  Phys} \textbf{2020}, \emph{128}, 171104\relax
\mciteBstWouldAddEndPuncttrue
\mciteSetBstMidEndSepPunct{\mcitedefaultmidpunct}
{\mcitedefaultendpunct}{\mcitedefaultseppunct}\relax
\EndOfBibitem
\bibitem[Kamal \latin{et~al.}(2021)Kamal, Tran, Kim, Wang, Chen, Cao, Joseph,
  and Ramprasad]{kamal_novel_2021}
Kamal,~D.; Tran,~H.; Kim,~C.; Wang,~Y.; Chen,~L.; Cao,~Y.; Joseph,~V.~R.;
  Ramprasad,~R. Novel high voltage polymer insulators using computational and
  data-driven techniques. \emph{Journal of Chemical Physics} \textbf{2021},
  \emph{154}, Publisher: AIP Publishing, LLC\relax
\mciteBstWouldAddEndPuncttrue
\mciteSetBstMidEndSepPunct{\mcitedefaultmidpunct}
{\mcitedefaultendpunct}{\mcitedefaultseppunct}\relax
\EndOfBibitem
\bibitem[Xu \latin{et~al.}(2023)Xu, Wang, and
  Barati~Farimani]{xu2023transpolymer}
Xu,~C.; Wang,~Y.; Barati~Farimani,~A. TransPolymer: a Transformer-based
  language model for polymer property predictions. \emph{npj Computational
  Materials} \textbf{2023}, \emph{9}, 64\relax
\mciteBstWouldAddEndPuncttrue
\mciteSetBstMidEndSepPunct{\mcitedefaultmidpunct}
{\mcitedefaultendpunct}{\mcitedefaultseppunct}\relax
\EndOfBibitem
\bibitem[Kuenneth and Ramprasad(2023)Kuenneth, and
  Ramprasad]{kuenneth_polybert_2023}
Kuenneth,~C.; Ramprasad,~R. {polyBERT}: a chemical language model to enable
  fully machine-driven ultrafast polymer informatics. \emph{Nature
  Communications} \textbf{2023}, \emph{14}, 4099, Number: 1 Publisher: Nature
  Publishing Group\relax
\mciteBstWouldAddEndPuncttrue
\mciteSetBstMidEndSepPunct{\mcitedefaultmidpunct}
{\mcitedefaultendpunct}{\mcitedefaultseppunct}\relax
\EndOfBibitem
\bibitem[Vaswani \latin{et~al.}(2017)Vaswani, Shazeer, Parmar, Uszkoreit,
  Jones, Gomez, Kaiser, and Polosukhin]{vaswani_attention_2017}
Vaswani,~A.; Shazeer,~N.; Parmar,~N.; Uszkoreit,~J.; Jones,~L.; Gomez,~A.~N.;
  Kaiser,~{\L}.; Polosukhin,~I. Attention is {All} you {Need}. Advances in
  {Neural} {Information} {Processing} {Systems}. 2017\relax
\mciteBstWouldAddEndPuncttrue
\mciteSetBstMidEndSepPunct{\mcitedefaultmidpunct}
{\mcitedefaultendpunct}{\mcitedefaultseppunct}\relax
\EndOfBibitem
\bibitem[Kotsiliti(2022)]{kotsiliti2022novo}
Kotsiliti,~E. De novo protein design with a language model. \emph{Nature
  Biotechnology} \textbf{2022}, \emph{40}, 1433--1433\relax
\mciteBstWouldAddEndPuncttrue
\mciteSetBstMidEndSepPunct{\mcitedefaultmidpunct}
{\mcitedefaultendpunct}{\mcitedefaultseppunct}\relax
\EndOfBibitem
\bibitem[Meier \latin{et~al.}(2021)Meier, Rao, Verkuil, Liu, Sercu, and
  Rives]{meier2021language}
Meier,~J.; Rao,~R.; Verkuil,~R.; Liu,~J.; Sercu,~T.; Rives,~A. Language models
  enable zero-shot prediction of the effects of mutations on protein function.
  \emph{Advances in Neural Information Processing Systems} \textbf{2021},
  \emph{34}, 29287--29303\relax
\mciteBstWouldAddEndPuncttrue
\mciteSetBstMidEndSepPunct{\mcitedefaultmidpunct}
{\mcitedefaultendpunct}{\mcitedefaultseppunct}\relax
\EndOfBibitem
\bibitem[Madani \latin{et~al.}(2023)Madani, Krause, Greene, Subramanian, Mohr,
  Holton, Olmos~Jr, Xiong, Sun, Socher, \latin{et~al.} others]{madani2023large}
Madani,~A.; Krause,~B.; Greene,~E.~R.; Subramanian,~S.; Mohr,~B.~P.;
  Holton,~J.~M.; Olmos~Jr,~J.~L.; Xiong,~C.; Sun,~Z.~Z.; Socher,~R.,
  \latin{et~al.}  Large language models generate functional protein sequences
  across diverse families. \emph{Nature Biotechnology} \textbf{2023},
  1--8\relax
\mciteBstWouldAddEndPuncttrue
\mciteSetBstMidEndSepPunct{\mcitedefaultmidpunct}
{\mcitedefaultendpunct}{\mcitedefaultseppunct}\relax
\EndOfBibitem
\bibitem[Ferruz \latin{et~al.}(2022)Ferruz, Schmidt, and
  H{\"o}cker]{ferruz2022protgpt2}
Ferruz,~N.; Schmidt,~S.; H{\"o}cker,~B. ProtGPT2 is a deep unsupervised
  language model for protein design. \emph{Nature communications}
  \textbf{2022}, \emph{13}, 4348\relax
\mciteBstWouldAddEndPuncttrue
\mciteSetBstMidEndSepPunct{\mcitedefaultmidpunct}
{\mcitedefaultendpunct}{\mcitedefaultseppunct}\relax
\EndOfBibitem
\bibitem[Ferruz and H{\"o}cker(2022)Ferruz, and
  H{\"o}cker]{ferruz2022controllable}
Ferruz,~N.; H{\"o}cker,~B. Controllable protein design with language models.
  \emph{Nature Machine Intelligence} \textbf{2022}, \emph{4}, 521--532\relax
\mciteBstWouldAddEndPuncttrue
\mciteSetBstMidEndSepPunct{\mcitedefaultmidpunct}
{\mcitedefaultendpunct}{\mcitedefaultseppunct}\relax
\EndOfBibitem
\bibitem[Brandes \latin{et~al.}(2022)Brandes, Ofer, Peleg, Rappoport, and
  Linial]{brandes2022proteinbert}
Brandes,~N.; Ofer,~D.; Peleg,~Y.; Rappoport,~N.; Linial,~M. ProteinBERT: a
  universal deep-learning model of protein sequence and function.
  \emph{Bioinformatics} \textbf{2022}, \emph{38}, 2102--2110\relax
\mciteBstWouldAddEndPuncttrue
\mciteSetBstMidEndSepPunct{\mcitedefaultmidpunct}
{\mcitedefaultendpunct}{\mcitedefaultseppunct}\relax
\EndOfBibitem
\bibitem[Chithrananda \latin{et~al.}(2020)Chithrananda, Grand, and
  Ramsundar]{chithrananda2020chemberta}
Chithrananda,~S.; Grand,~G.; Ramsundar,~B. ChemBERTa: Large-Scale
  Self-Supervised Pretraining for Molecular Property Prediction. \textbf{2020},
  \relax
\mciteBstWouldAddEndPunctfalse
\mciteSetBstMidEndSepPunct{\mcitedefaultmidpunct}
{}{\mcitedefaultseppunct}\relax
\EndOfBibitem
\bibitem[Blanchard \latin{et~al.}(2022)Blanchard, Chandra~Shekar, Gao, Gounley,
  Lyngaas, Glaser, and Bhowmik]{Blanchard2022}
Blanchard,~A.~E.; Chandra~Shekar,~M.; Gao,~S.; Gounley,~J.; Lyngaas,~I.;
  Glaser,~J.; Bhowmik,~D. {Automating Genetic Algorithm Mutations for Molecules
  Using a Masked Language Model}. \emph{IEEE Transactions on Evolutionary
  Computation} \textbf{2022}, \relax
\mciteBstWouldAddEndPunctfalse
\mciteSetBstMidEndSepPunct{\mcitedefaultmidpunct}
{}{\mcitedefaultseppunct}\relax
\EndOfBibitem
\bibitem[Blanchard \latin{et~al.}(2023)Blanchard, Bhowmik, Fox, Gounley,
  Glaser, Akpa, and Irle]{blanchard2023adaptive}
Blanchard,~A.~E.; Bhowmik,~D.; Fox,~Z.; Gounley,~J.; Glaser,~J.; Akpa,~B.~S.;
  Irle,~S. Adaptive language model training for molecular design. \emph{Journal
  of Cheminformatics} \textbf{2023}, \emph{15}, 1--12\relax
\mciteBstWouldAddEndPuncttrue
\mciteSetBstMidEndSepPunct{\mcitedefaultmidpunct}
{\mcitedefaultendpunct}{\mcitedefaultseppunct}\relax
\EndOfBibitem
\bibitem[Ross \latin{et~al.}(2022)Ross, Belgodere, Chenthamarakshan, Padhi,
  Mroueh, and Das]{ross2022large}
Ross,~J.; Belgodere,~B.; Chenthamarakshan,~V.; Padhi,~I.; Mroueh,~Y.; Das,~P.
  Large-scale chemical language representations capture molecular structure and
  properties. \emph{Nature Machine Intelligence} \textbf{2022}, \emph{4},
  1256--1264\relax
\mciteBstWouldAddEndPuncttrue
\mciteSetBstMidEndSepPunct{\mcitedefaultmidpunct}
{\mcitedefaultendpunct}{\mcitedefaultseppunct}\relax
\EndOfBibitem
\bibitem[Ahmad \latin{et~al.}(2022)Ahmad, Simon, Chithrananda, Grand, and
  Ramsundar]{ahmad2022chemberta}
Ahmad,~W.; Simon,~E.; Chithrananda,~S.; Grand,~G.; Ramsundar,~B. Chemberta-2:
  Towards chemical foundation models. \emph{arXiv preprint arXiv:2209.01712}
  \textbf{2022}, \relax
\mciteBstWouldAddEndPunctfalse
\mciteSetBstMidEndSepPunct{\mcitedefaultmidpunct}
{}{\mcitedefaultseppunct}\relax
\EndOfBibitem
\bibitem[Lipman and Pearson(1985)Lipman, and Pearson]{lipman1985rapid}
Lipman,~D.~J.; Pearson,~W.~R. Rapid and sensitive protein similarity searches.
  \emph{Science} \textbf{1985}, \emph{227}, 1435--1441\relax
\mciteBstWouldAddEndPuncttrue
\mciteSetBstMidEndSepPunct{\mcitedefaultmidpunct}
{\mcitedefaultendpunct}{\mcitedefaultseppunct}\relax
\EndOfBibitem
\bibitem[Shivanyuk \latin{et~al.}(2007)Shivanyuk, Ryabukhin, Tolmachev,
  Bogolyubsky, Mykytenko, Chupryna, Heilman, and Kostyuk]{shivanyuk2007enamine}
Shivanyuk,~A.; Ryabukhin,~S.; Tolmachev,~A.; Bogolyubsky,~A.; Mykytenko,~D.;
  Chupryna,~A.; Heilman,~W.; Kostyuk,~A. Enamine real database: Making chemical
  diversity real. \emph{Chemistry today} \textbf{2007}, \emph{25}, 58--59\relax
\mciteBstWouldAddEndPuncttrue
\mciteSetBstMidEndSepPunct{\mcitedefaultmidpunct}
{\mcitedefaultendpunct}{\mcitedefaultseppunct}\relax
\EndOfBibitem
\bibitem[Kim \latin{et~al.}(2021)Kim, Chen, Cheng, Gindulyte, He, He, Li,
  Shoemaker, Thiessen, Yu, \latin{et~al.} others]{kim2021pubchem}
Kim,~S.; Chen,~J.; Cheng,~T.; Gindulyte,~A.; He,~J.; He,~S.; Li,~Q.;
  Shoemaker,~B.~A.; Thiessen,~P.~A.; Yu,~B., \latin{et~al.}  PubChem in 2021:
  new data content and improved web interfaces. \emph{Nucleic acids research}
  \textbf{2021}, \emph{49}, D1388--D1395\relax
\mciteBstWouldAddEndPuncttrue
\mciteSetBstMidEndSepPunct{\mcitedefaultmidpunct}
{\mcitedefaultendpunct}{\mcitedefaultseppunct}\relax
\EndOfBibitem
\bibitem[Irwin and Shoichet(2005)Irwin, and Shoichet]{irwin2005zinc}
Irwin,~J.~J.; Shoichet,~B.~K. ZINC- a free database of commercially available
  compounds for virtual screening. \emph{Journal of chemical information and
  modeling} \textbf{2005}, \emph{45}, 177--182\relax
\mciteBstWouldAddEndPuncttrue
\mciteSetBstMidEndSepPunct{\mcitedefaultmidpunct}
{\mcitedefaultendpunct}{\mcitedefaultseppunct}\relax
\EndOfBibitem
\bibitem[Consortium(2019)]{uniprot2019uniprot}
Consortium,~U. UniProt: a worldwide hub of protein knowledge. \emph{Nucleic
  acids research} \textbf{2019}, \emph{47}, D506--D515\relax
\mciteBstWouldAddEndPuncttrue
\mciteSetBstMidEndSepPunct{\mcitedefaultmidpunct}
{\mcitedefaultendpunct}{\mcitedefaultseppunct}\relax
\EndOfBibitem
\bibitem[Kuenneth \latin{et~al.}(2021)Kuenneth, Rajan, Tran, Chen, Kim, and
  Ramprasad]{kuenneth2021polymer}
Kuenneth,~C.; Rajan,~A.~C.; Tran,~H.; Chen,~L.; Kim,~C.; Ramprasad,~R. Polymer
  informatics with multi-task learning. \emph{Patterns} \textbf{2021},
  \emph{2}, 100238\relax
\mciteBstWouldAddEndPuncttrue
\mciteSetBstMidEndSepPunct{\mcitedefaultmidpunct}
{\mcitedefaultendpunct}{\mcitedefaultseppunct}\relax
\EndOfBibitem
\bibitem[Otsuka \latin{et~al.}(2011)Otsuka, Kuwajima, Hosoya, Xu, and
  Yamazaki]{otsuka2011polyinfo}
Otsuka,~S.; Kuwajima,~I.; Hosoya,~J.; Xu,~Y.; Yamazaki,~M. PoLyInfo: Polymer
  database for polymeric materials design. 2011 International Conference on
  Emerging Intelligent Data and Web Technologies. 2011; pp 22--29\relax
\mciteBstWouldAddEndPuncttrue
\mciteSetBstMidEndSepPunct{\mcitedefaultmidpunct}
{\mcitedefaultendpunct}{\mcitedefaultseppunct}\relax
\EndOfBibitem
\bibitem[Kim \latin{et~al.}(2018)Kim, Chandrasekaran, Huan, Das, and
  Ramprasad]{kim2018polymer}
Kim,~C.; Chandrasekaran,~A.; Huan,~T.~D.; Das,~D.; Ramprasad,~R. Polymer
  genome: a data-powered polymer informatics platform for property predictions.
  \emph{The Journal of Physical Chemistry C} \textbf{2018}, \emph{122},
  17575--17585\relax
\mciteBstWouldAddEndPuncttrue
\mciteSetBstMidEndSepPunct{\mcitedefaultmidpunct}
{\mcitedefaultendpunct}{\mcitedefaultseppunct}\relax
\EndOfBibitem
\bibitem[Ma and Luo(2020)Ma, and Luo]{ma_pi1m_2020}
Ma,~R.; Luo,~T. {PI1M}: {A} {Benchmark} {Database} for {Polymer} {Informatics}.
  \emph{Cite This: J. Chem. Inf. Model} \textbf{2020}, \emph{2020},
  4684--4690\relax
\mciteBstWouldAddEndPuncttrue
\mciteSetBstMidEndSepPunct{\mcitedefaultmidpunct}
{\mcitedefaultendpunct}{\mcitedefaultseppunct}\relax
\EndOfBibitem
\bibitem[Devlin \latin{et~al.}(2018)Devlin, Chang, Lee, and
  Toutanova]{devlin2018bert}
Devlin,~J.; Chang,~M.-W.; Lee,~K.; Toutanova,~K. Bert: Pre-training of deep
  bidirectional transformers for language understanding. \emph{arXiv preprint
  arXiv:1810.04805} \textbf{2018}, \relax
\mciteBstWouldAddEndPunctfalse
\mciteSetBstMidEndSepPunct{\mcitedefaultmidpunct}
{}{\mcitedefaultseppunct}\relax
\EndOfBibitem
\bibitem[Wolf \latin{et~al.}(2020)Wolf, Debut, Sanh, Chaumond, Delangue, Moi,
  Cistac, Rault, Louf, Funtowicz, Davison, Shleifer, von Platen, Ma, Jernite,
  Plu, Xu, Le~Scao, Gugger, Drame, Lhoest, and
  Rush]{wolf-etal-2020-transformers}
Wolf,~T. \latin{et~al.}  Transformers: State-of-the-Art Natural Language
  Processing. Proceedings of the 2020 Conference on Empirical Methods in
  Natural Language Processing: System Demonstrations. Online, 2020; pp
  38--45\relax
\mciteBstWouldAddEndPuncttrue
\mciteSetBstMidEndSepPunct{\mcitedefaultmidpunct}
{\mcitedefaultendpunct}{\mcitedefaultseppunct}\relax
\EndOfBibitem
\bibitem[Gurnani \latin{et~al.}(2023)Gurnani, Kuenneth, Toland, and
  Ramprasad]{gurnani2023polymer}
Gurnani,~R.; Kuenneth,~C.; Toland,~A.; Ramprasad,~R. Polymer Informatics at
  Scale with Multitask Graph Neural Networks. \emph{Chemistry of Materials}
  \textbf{2023}, \emph{35}, 1560--1567\relax
\mciteBstWouldAddEndPuncttrue
\mciteSetBstMidEndSepPunct{\mcitedefaultmidpunct}
{\mcitedefaultendpunct}{\mcitedefaultseppunct}\relax
\EndOfBibitem
\bibitem[Acharya \latin{et~al.}(2020)Acharya, Agarwal, Baker, Baudry, Bhowmik,
  Boehm, Byler, Chen, Coates, Cooper, \latin{et~al.}
  others]{acharya2020supercomputer}
Acharya,~A.; Agarwal,~R.; Baker,~M.~B.; Baudry,~J.; Bhowmik,~D.; Boehm,~S.;
  Byler,~K.~G.; Chen,~S.; Coates,~L.; Cooper,~C.~J., \latin{et~al.}
  Supercomputer-based ensemble docking drug discovery pipeline with application
  to COVID-19. \emph{Journal of chemical information and modeling}
  \textbf{2020}, \emph{60}, 5832--5852\relax
\mciteBstWouldAddEndPuncttrue
\mciteSetBstMidEndSepPunct{\mcitedefaultmidpunct}
{\mcitedefaultendpunct}{\mcitedefaultseppunct}\relax
\EndOfBibitem
\bibitem[Blanchard \latin{et~al.}(2022)Blanchard, Gounley, Bhowmik, Shekar,
  Lyngaas, Gao, Yin, Tsaris, Wang, and Glaser]{blanchard_SARS-CoV-2-2022}
Blanchard,~A.~E.; Gounley,~J.; Bhowmik,~D.; Shekar,~M.~C.; Lyngaas,~I.;
  Gao,~S.; Yin,~J.; Tsaris,~A.; Wang,~F.; Glaser,~J. Language models for the
  prediction of SARS-CoV-2 inhibitors. \emph{The International Journal of High
  Performance Computing Applications} \textbf{2022}, \emph{36}, 587--602\relax
\mciteBstWouldAddEndPuncttrue
\mciteSetBstMidEndSepPunct{\mcitedefaultmidpunct}
{\mcitedefaultendpunct}{\mcitedefaultseppunct}\relax
\EndOfBibitem
\bibitem[Yang \latin{et~al.}(2022)Yang, Tao, He, McCutcheon, and
  Li]{yang2022machine}
Yang,~J.; Tao,~L.; He,~J.; McCutcheon,~J.~R.; Li,~Y. Machine learning enables
  interpretable discovery of innovative polymers for gas separation membranes.
  \emph{Science Advances} \textbf{2022}, \emph{8}, eabn9545\relax
\mciteBstWouldAddEndPuncttrue
\mciteSetBstMidEndSepPunct{\mcitedefaultmidpunct}
{\mcitedefaultendpunct}{\mcitedefaultseppunct}\relax
\EndOfBibitem
\bibitem[Rajbhandari \latin{et~al.}(2020)Rajbhandari, Rasley, Ruwase, and
  He]{Rajbhandari_deepspeed_2020}
Rajbhandari,~S.; Rasley,~J.; Ruwase,~O.; He,~Y. ZeRO: Memory Optimizations
  toward Training Trillion Parameter Models. Proceedings of the International
  Conference for High Performance Computing, Networking, Storage and Analysis.
  2020\relax
\mciteBstWouldAddEndPuncttrue
\mciteSetBstMidEndSepPunct{\mcitedefaultmidpunct}
{\mcitedefaultendpunct}{\mcitedefaultseppunct}\relax
\EndOfBibitem
\bibitem[You \latin{et~al.}(2019)You, Li, Reddi, Hseu, Kumar, Bhojanapalli,
  Song, Demmel, Keutzer, and Hsieh]{you2019large}
You,~Y.; Li,~J.; Reddi,~S.; Hseu,~J.; Kumar,~S.; Bhojanapalli,~S.; Song,~X.;
  Demmel,~J.; Keutzer,~K.; Hsieh,~C.-J. Large batch optimization for deep
  learning: Training bert in 76 minutes. \emph{arXiv preprint arXiv:1904.00962}
  \textbf{2019}, \relax
\mciteBstWouldAddEndPunctfalse
\mciteSetBstMidEndSepPunct{\mcitedefaultmidpunct}
{}{\mcitedefaultseppunct}\relax
\EndOfBibitem
\bibitem[Abnar \latin{et~al.}(2021)Abnar, Dehghani, Neyshabur, and
  Sedghi]{abnar2021exploring}
Abnar,~S.; Dehghani,~M.; Neyshabur,~B.; Sedghi,~H. Exploring the limits of
  large scale pre-training. \emph{arXiv preprint arXiv:2110.02095}
  \textbf{2021}, \relax
\mciteBstWouldAddEndPunctfalse
\mciteSetBstMidEndSepPunct{\mcitedefaultmidpunct}
{}{\mcitedefaultseppunct}\relax
\EndOfBibitem
\end{mcitethebibliography}
\providecommand{\latin}[1]{#1}
\makeatletter
\providecommand{\doi}
  {\begingroup\let\do\@makeother\dospecials
  \catcode`\{=1 \catcode`\}=2 \doi@aux}
\providecommand{\doi@aux}[1]{\endgroup\texttt{#1}}
\makeatother
\providecommand*\mcitethebibliography{\thebibliography}
\csname @ifundefined\endcsname{endmcitethebibliography}
  {\let\endmcitethebibliography\endthebibliography}{}

\newpage
\setcounter{figure}{0}
\renewcommand\thefigure{S.\arabic{figure}}    
\setcounter{table}{0}
\renewcommand{\thetable}{S.\arabic{table}}

\begin{figure}[ht]
\centering
\includegraphics[width=0.8\linewidth]{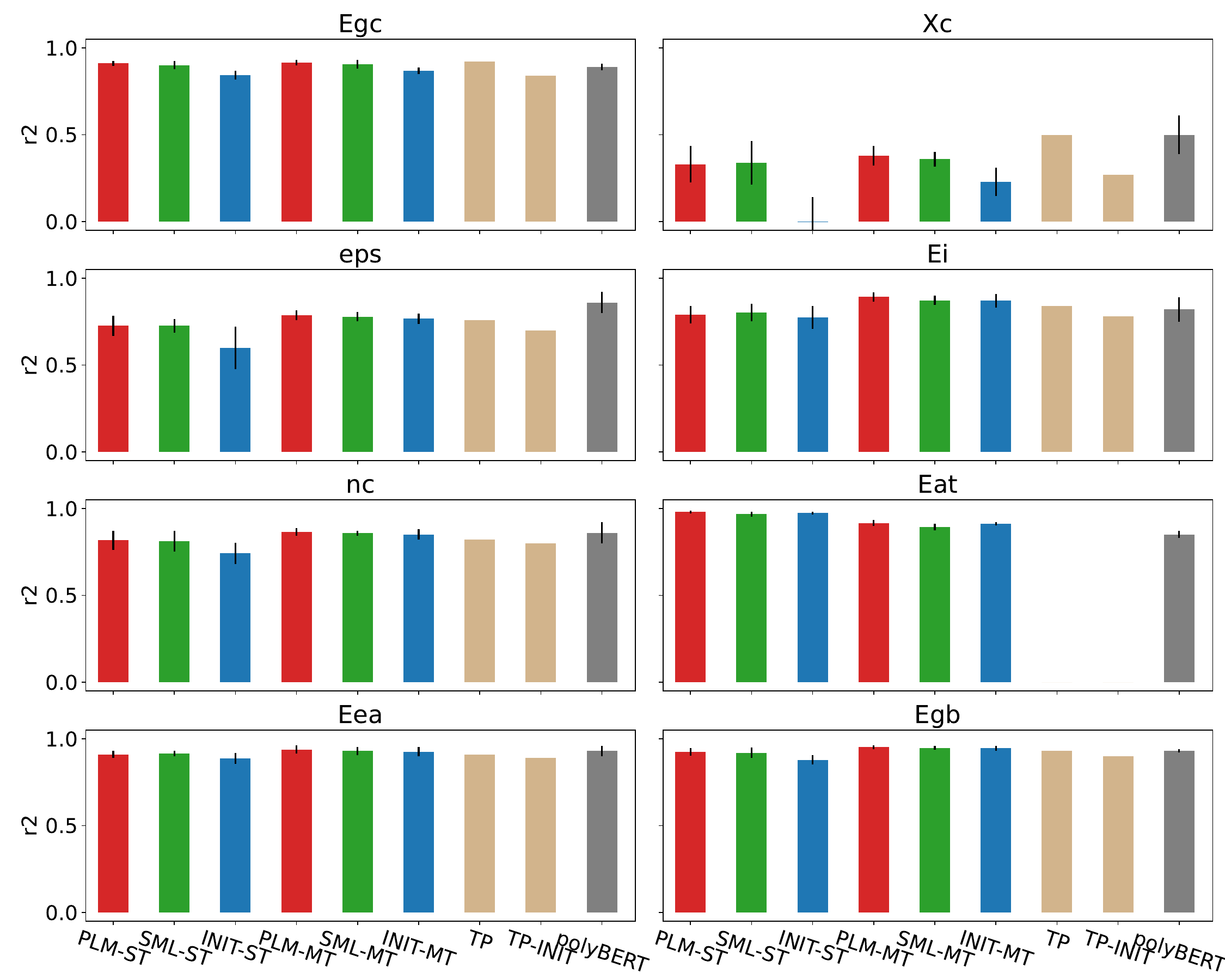}
\caption{R$^2$ prediction accuracy of DFT properties. Error bars show the STD from the 5-fold cross-validation. Results of TransPolymer \cite{xu2023transpolymer} (no STD reported, hence no error bars plotted), where TP and TP-INIT represent the models with and without pre-training, respectively, and polyBERT \cite{kuenneth_polybert_2023} are also shown for reference.
}
\label{fig:bar-R2}
\end{figure}

\begin{table}[h]
  \begin{center}
    \caption{Prediction error: mean relative RMSE together with the STD from the 5-fold cross-validation of fine-tuned models on DFT datasets}
    \label{tab:finetuning-RelativeRMSE}
    \resizebox{\textwidth}{!}{
    \begin{tabular}{l|c|c|c|c|c|c|c|c|c} 
\textbf{Model}&&Egc&Xc&eps&Ei&nc&Eat&Eea&Egb\\\hline
\multirow{ 2}{*}{PLM-ST} & train&0.010$\pm$ 0.0010 & 0.024$\pm$ 0.0040 & 0.017$\pm$ 0.0028 & 0.015$\pm$ 0.0018 & 0.010$\pm$ 0.0013 & 0.010$\pm$ 0.0021 & 0.018$\pm$ 0.0020 & 0.015$\pm$ 0.0014\\
 & test&0.046$\pm$ 0.0037 & 0.192$\pm$ 0.0131 & 0.096$\pm$ 0.0100 & 0.074$\pm$ 0.0104 & 0.050$\pm$ 0.0053 & \textbf{0.025$\pm$ 0.0053} & 0.070$\pm$ 0.0082 & 0.055$\pm$ 0.0085\\\hline
\multirow{ 2}{*}{SML-ST} & train&0.011$\pm$ 0.0016 & 0.024$\pm$ 0.0043 & 0.019$\pm$ 0.0024 & 0.017$\pm$ 0.0011 & 0.011$\pm$ 0.0013 & 0.014$\pm$ 0.0019 & 0.022$\pm$ 0.0037 & 0.017$\pm$ 0.0018\\
 & test&0.049$\pm$ 0.0057 & 0.190$\pm$ 0.0126 & 0.096$\pm$ 0.0034 & 0.072$\pm$ 0.0101 & 0.051$\pm$ 0.0050 & 0.031$\pm$ 0.0073 & 0.068$\pm$ 0.0035 & 0.057$\pm$ 0.0115\\\hline
\multirow{ 2}{*}{INIT-ST} & train&0.011$\pm$ 0.0005 & 0.025$\pm$ 0.0013 & 0.017$\pm$ 0.0019 & 0.015$\pm$ 0.0038 & 0.011$\pm$ 0.0025 & 0.010$\pm$ 0.0025 & 0.021$\pm$ 0.0051 & 0.017$\pm$ 0.0037\\
 & test&0.062$\pm$ 0.0054 & 0.236$\pm$ 0.0197 & 0.115$\pm$ 0.0112 & 0.076$\pm$ 0.0121 & 0.060$\pm$ 0.0043 & \textbf{0.029$\pm$ 0.0036} & 0.077$\pm$ 0.0067 & 0.070$\pm$ 0.0082\\\hline
\multirow{ 2}{*}{PLM-MT} & train&0.011$\pm$ 0.0012 & 0.024$\pm$ 0.0019 & 0.017$\pm$ 0.0018 & 0.015$\pm$ 0.0015 & 0.009$\pm$ 0.0011 & 0.014$\pm$ 0.0013 & 0.020$\pm$ 0.0010 & 0.014$\pm$ 0.0015\\
 & test&\textbf{0.045$\pm$ 0.0045} & 0.185$\pm$ 0.0082 & \textbf{0.085$\pm$ 0.0079} & \textbf{0.053$\pm$ 0.0096} & \textbf{0.043$\pm$ 0.0032} & 0.051$\pm$ 0.0070 & \textbf{0.057$\pm$ 0.0076} & \textbf{0.044$\pm$ 0.0049}\\\hline
\multirow{ 2}{*}{SML-MT} & train&0.013$\pm$ 0.0007 & 0.027$\pm$ 0.0017 & 0.019$\pm$ 0.0032 & 0.016$\pm$ 0.0012 & 0.012$\pm$ 0.0016 & 0.018$\pm$ 0.0010 & 0.025$\pm$ 0.0020 & 0.017$\pm$ 0.0014\\
 & test&0.048$\pm$ 0.0060 & 0.189$\pm$ 0.0134 & 0.087$\pm$ 0.0090 & \textbf{0.058$\pm$ 0.0100} & \textbf{0.045$\pm$ 0.0027} & 0.058$\pm$ 0.0072 & \textbf{0.060$\pm$ 0.0065} & \textbf{0.046$\pm$ 0.0049}\\\hline
\multirow{ 2}{*}{INIT-MT} & train&0.010$\pm$ 0.0008 & 0.021$\pm$ 0.0034 & 0.013$\pm$ 0.0014 & 0.012$\pm$ 0.0014 & 0.009$\pm$ 0.0007 & 0.014$\pm$ 0.0022 & 0.017$\pm$ 0.0026 & 0.013$\pm$ 0.0023\\
 & test&0.057$\pm$ 0.0041 & 0.207$\pm$ 0.0176 & 0.089$\pm$ 0.0053 & \textbf{0.058$\pm$ 0.0104} & 0.046$\pm$ 0.0036 & 0.053$\pm$ 0.0050 & 0.062$\pm$ 0.0087 & 0.047$\pm$ 0.0059\\\hline
TransPolymer & test& \textbf{0.044} & \textbf{0.166} & 0.087 & 0.065 & 0.050 & - & 0.070 & 0.054\\\hline
TransPolymer-INIT & test&0.063 & 0.201 & 0.098 & 0.077 & 0.050 & - & 0.078 & 0.064\\\hline
polyBERT & test&0.048$\pm$ 0.0040 & \textbf{0.167$\pm$ 0.0255} & \textbf{0.070$\pm$ 0.0183} & 0.070$\pm$ 0.0100 & \textbf{0.045$\pm$ 0.0150} & 0.065$\pm$ 0.0050 & 0.061$\pm$ 0.0087 & 0.051$\pm$ 0.0042\\\hline
 \end{tabular}}
  \end{center}
\end{table}

\begin{table}[h]
  \begin{center}
    \caption{Prediction accuracy: mean R$^2$ together with the STD from the 5-fold cross-validation of fine-tuned models on DFT datasets}
    \label{tab:finetuning-R2}
    \resizebox{\textwidth}{!}{
    \begin{tabular}{l|c|c|c|c|c|c|c|c|c} 
\textbf{Model}&&Egc&Xc&eps&Ei&nc&Eat&Eea&Egb\\\hline
\multirow{ 2}{*}{PLM-ST} & train&0.996$\pm$ 0.0008 & 0.989$\pm$ 0.0034 & 0.992$\pm$ 0.0030 & 0.992$\pm$ 0.0018 & 0.993$\pm$ 0.0021 & 0.997$\pm$ 0.0013 & 0.994$\pm$ 0.0012 & 0.995$\pm$ 0.0010\\
 & test&0.911$\pm$ 0.0141 & 0.330$\pm$ 0.1051 & 0.726$\pm$ 0.0587 & 0.791$\pm$ 0.0497 & 0.817$\pm$ 0.0560 & 0.980$\pm$ 0.0085 & 0.910$\pm$ 0.0198 & 0.925$\pm$ 0.0210\\\hline
\multirow{ 2}{*}{SML-ST} & train&0.995$\pm$ 0.0013 & 0.989$\pm$ 0.0037 & 0.989$\pm$ 0.0027 & 0.989$\pm$ 0.0011 & 0.991$\pm$ 0.0022 & 0.994$\pm$ 0.0016 & 0.991$\pm$ 0.0030 & 0.993$\pm$ 0.0015\\
 & test&0.901$\pm$ 0.0222 & 0.340$\pm$ 0.1257 & 0.726$\pm$ 0.0383 & 0.802$\pm$ 0.0511 & 0.812$\pm$ 0.0581 & 0.967$\pm$ 0.0150 & 0.915$\pm$ 0.0153 & 0.920$\pm$ 0.0290\\\hline
\multirow{ 2}{*}{INIT-ST} & train&0.995$\pm$ 0.0004 & 0.989$\pm$ 0.0011 & 0.991$\pm$ 0.0018 & 0.991$\pm$ 0.0045 & 0.991$\pm$ 0.0036 & 0.997$\pm$ 0.0017 & 0.992$\pm$ 0.0037 & 0.992$\pm$ 0.0032\\
 & test&0.843$\pm$ 0.0262 & -0.002$\pm$ 0.1420 & 0.598$\pm$ 0.1218 & 0.775$\pm$ 0.0649 & 0.741$\pm$ 0.0597 & 0.974$\pm$ 0.0068 & 0.888$\pm$ 0.0311 & 0.879$\pm$ 0.0269\\\hline
\multirow{ 2}{*}{PLM-MT} & train&0.995$\pm$ 0.0011 & 0.989$\pm$ 0.0018 & 0.992$\pm$ 0.0014 & 0.992$\pm$ 0.0018 & 0.994$\pm$ 0.0013 & 0.994$\pm$ 0.0012 & 0.993$\pm$ 0.0009 & 0.995$\pm$ 0.0011\\
 & test&0.916$\pm$ 0.0169 & 0.380$\pm$ 0.0572 & 0.787$\pm$ 0.0282 & 0.892$\pm$ 0.0272 & 0.866$\pm$ 0.0222 & 0.916$\pm$ 0.0175 & 0.938$\pm$ 0.0229 & 0.951$\pm$ 0.0108\\\hline
\multirow{ 2}{*}{SML-MT} & train&0.993$\pm$ 0.0008 & 0.987$\pm$ 0.0016 & 0.989$\pm$ 0.0039 & 0.991$\pm$ 0.0009 & 0.989$\pm$ 0.0028 & 0.990$\pm$ 0.0013 & 0.989$\pm$ 0.0017 & 0.993$\pm$ 0.0012\\
 & test&0.906$\pm$ 0.0245 & 0.360$\pm$ 0.0421 & 0.778$\pm$ 0.0264 & 0.872$\pm$ 0.0263 & 0.858$\pm$ 0.0141 & 0.893$\pm$ 0.0183 & 0.930$\pm$ 0.0240 & 0.947$\pm$ 0.0102\\\hline
\multirow{ 2}{*}{INIT-MT} & train&0.996$\pm$ 0.0007 & 0.992$\pm$ 0.0025 & 0.995$\pm$ 0.0010 & 0.995$\pm$ 0.0013 & 0.995$\pm$ 0.0009 & 0.994$\pm$ 0.0017 & 0.995$\pm$ 0.0017 & 0.996$\pm$ 0.0015\\
 & test&0.867$\pm$ 0.0191 & 0.229$\pm$ 0.0824 & 0.767$\pm$ 0.0295 & 0.870$\pm$ 0.0381 & 0.851$\pm$ 0.0293 & 0.911$\pm$ 0.0096 & 0.926$\pm$ 0.0261 & 0.946$\pm$ 0.0142\\\hline
TransPolymer & test&0.920 & 0.500 & 0.760 & 0.840 & 0.820 & - & 0.910 & 0.930\\\hline
TransPolymer-INIT & test&0.840 & 0.270 & 0.700 & 0.780 & 0.800 & - & 0.890 & 0.900\\\hline
polyBERT & test&0.890$\pm$ 0.0200 & 0.500$\pm$ 0.1100 & 0.860$\pm$ 0.0600 & 0.820$\pm$ 0.0700 & 0.860$\pm$ 0.0600 & 0.850$\pm$ 0.0200 & 0.930$\pm$ 0.0300 & 0.930$\pm$ 0.0100\\\hline
    \end{tabular}}
  \end{center}
\end{table}

\begin{figure}[ht]
\centering
\includegraphics[width=0.99\linewidth]{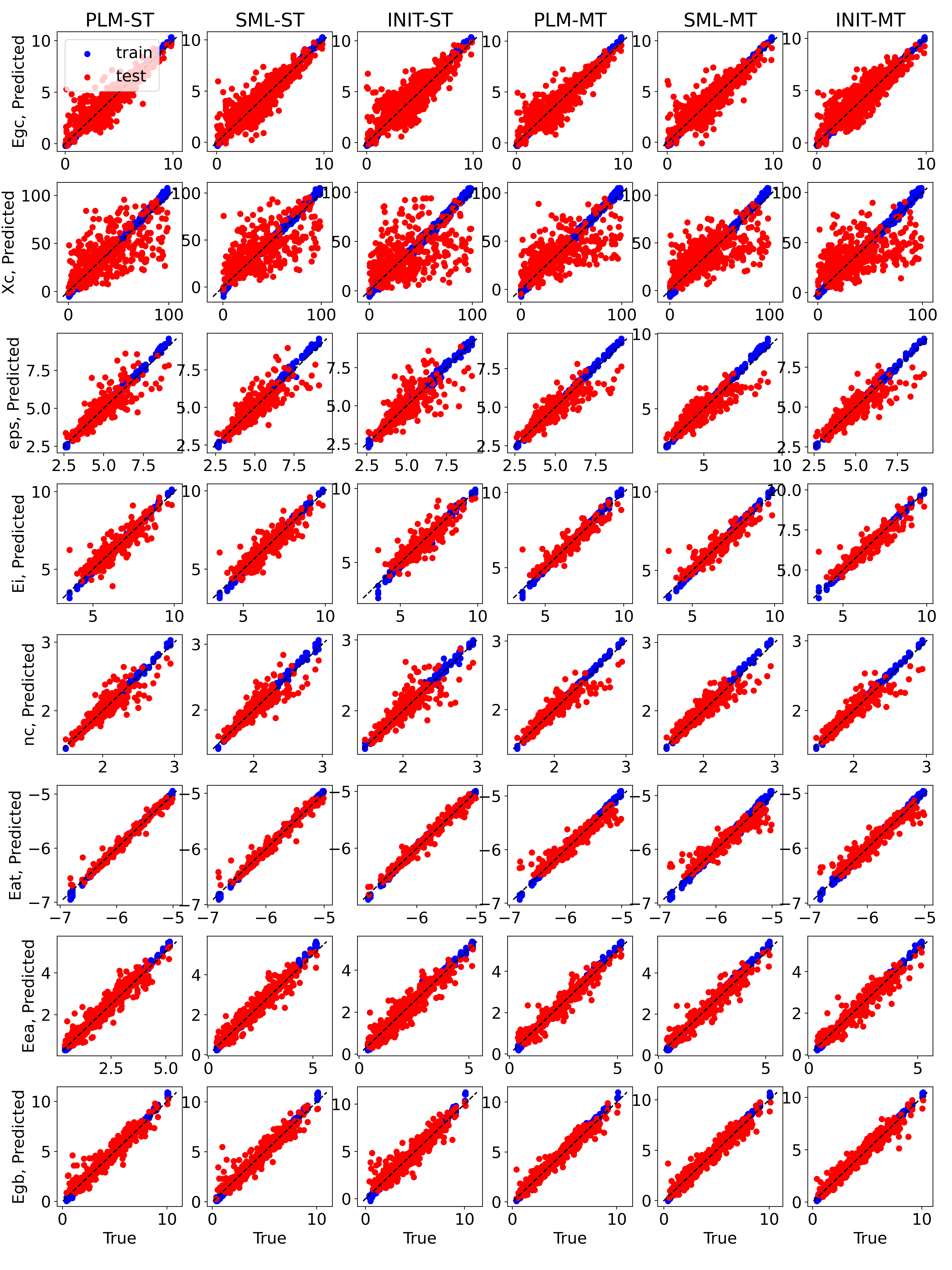}
\caption{Parity plots of 8 DFT properties for both training and test sets of all fine-tuned models from the 5-fold cross-validation.}
\label{fig:parityplot}
\end{figure}

\clearpage
\newpage
\begin{figure}[ht]
\centering
\subfigure[]{\includegraphics[width=1.0\textwidth]{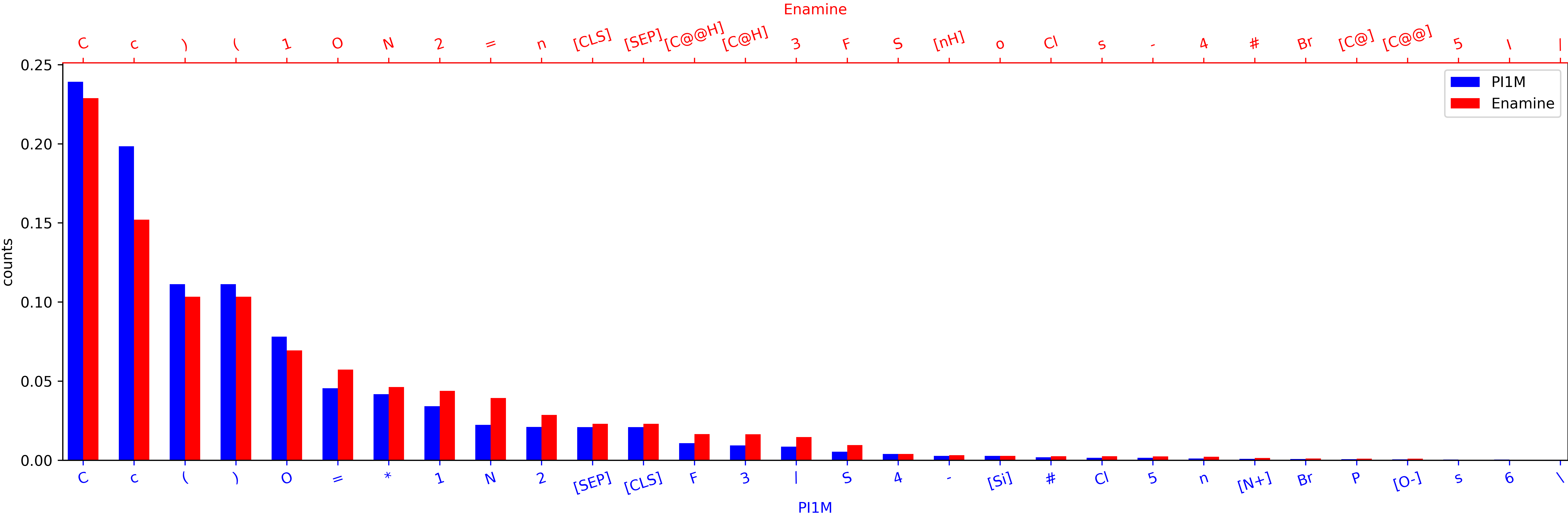}} 
\subfigure[]{\includegraphics[width=1.0\textwidth]{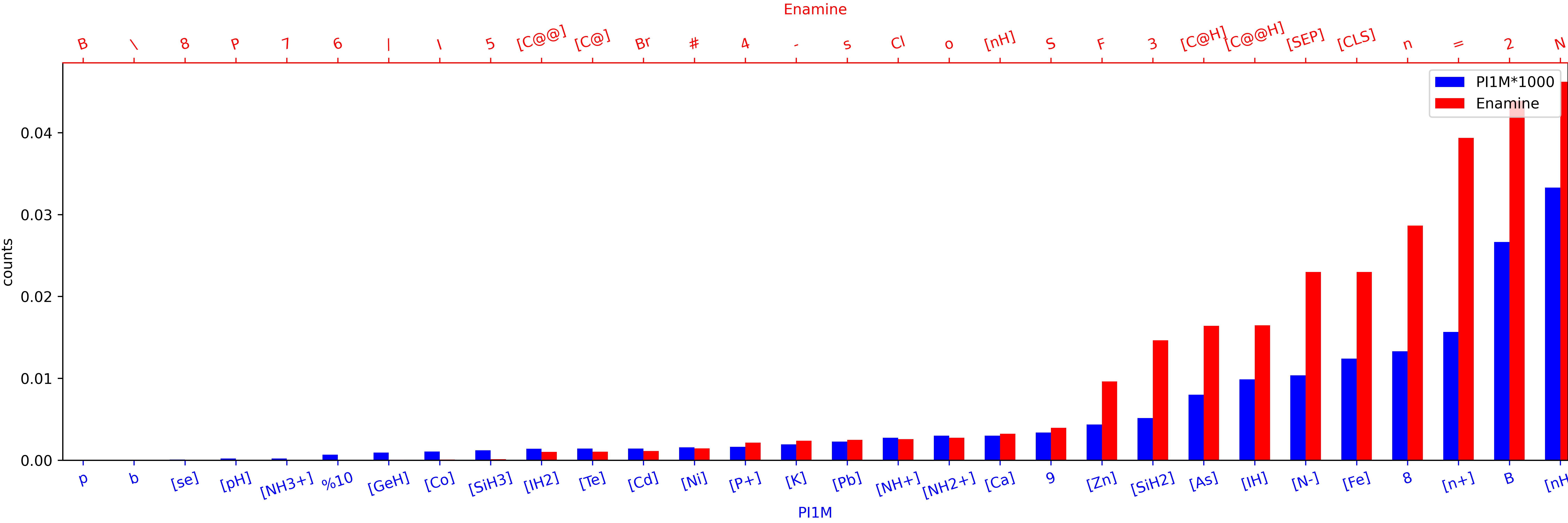}} 
\caption{Counts (percentage) of tokens in the two PI1M and Enamine REAL training sets: (a) top-20 most frequent and (b) top-20 least frequent (0s ignored).}
\label{fig:tokens-counts}
\end{figure}

\begin{figure}[ht]
\centering
\subfigure[]{\includegraphics[width=0.8\textwidth]{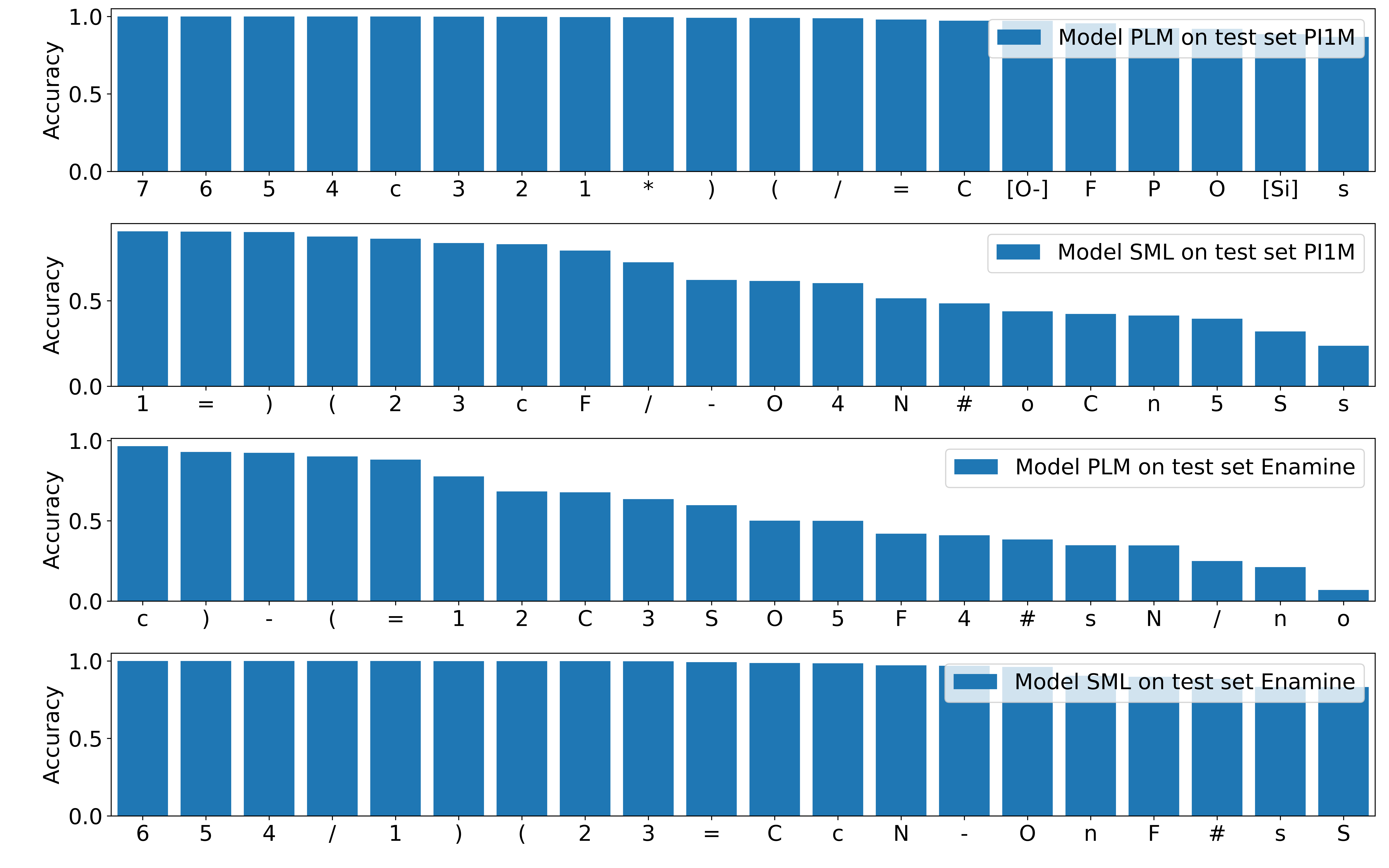}} 
\subfigure[]{\includegraphics[width=0.8\textwidth]{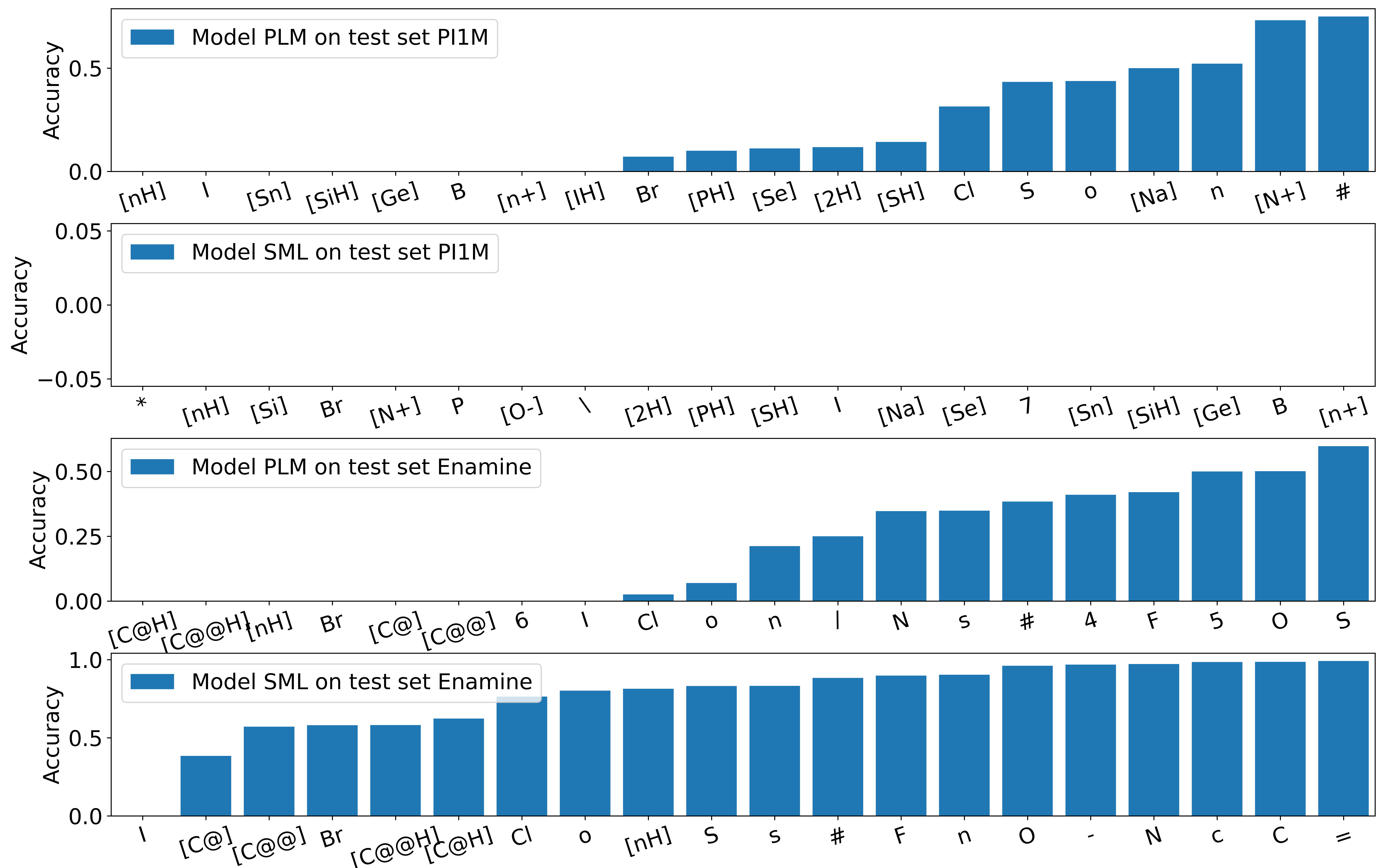}} 
\caption{Prediction accuracy of the (a) top-20 most accurately and (b) top-20 least accurately predicted tokens.}
\label{fig:tokens-accuracy}
\end{figure}

\begin{figure}[ht]
\centering
\includegraphics[width=0.95\linewidth]{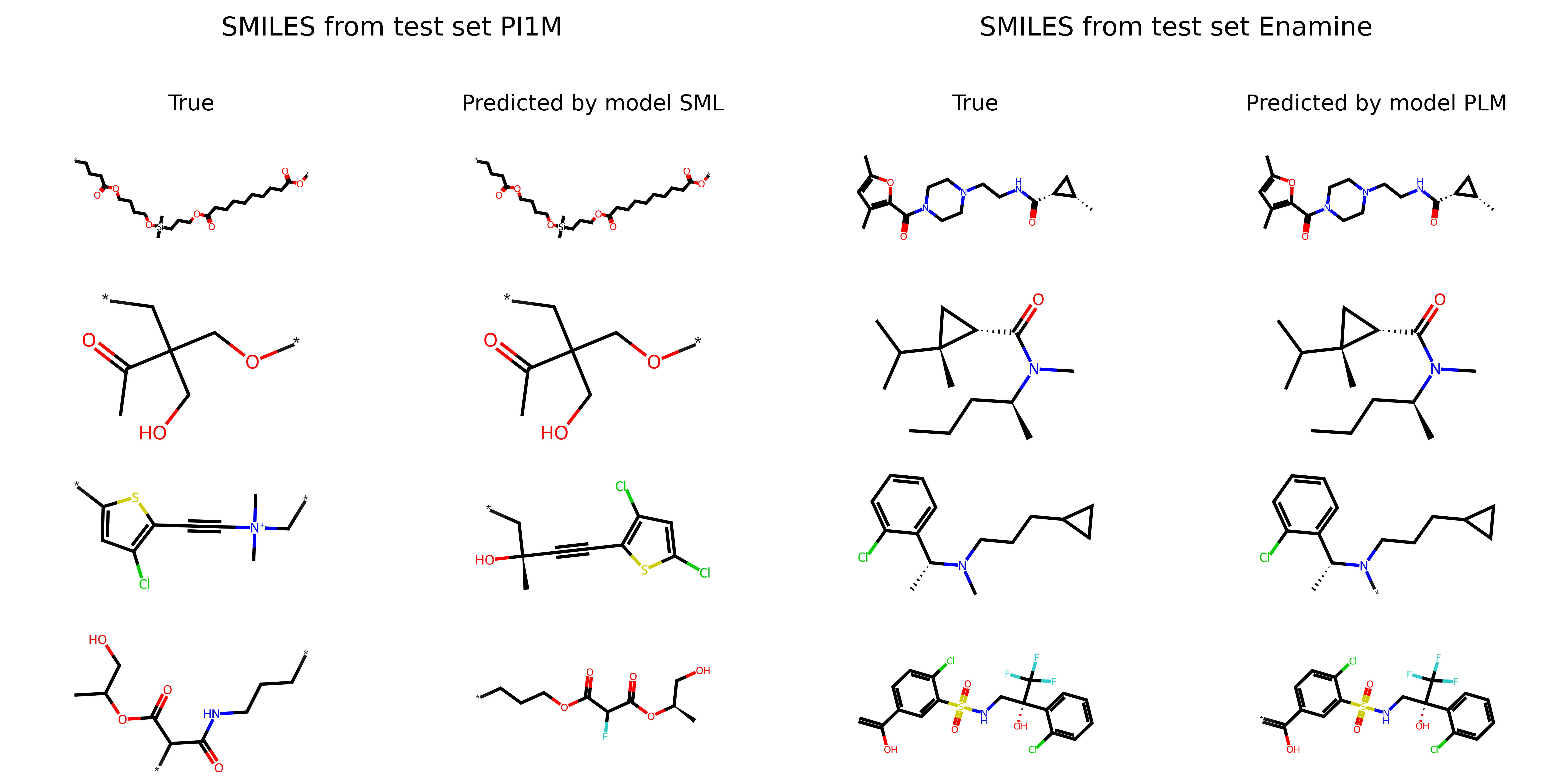}
\caption{Two successful (top-two rows) and two failed (bottom-two rows) SMILES in the out-of-distribution testing.}
\label{fig:SMILES-crossing}
\end{figure}
\clearpage
\newpage

\begin{figure}[ht]
\centering
\includegraphics[width=0.8\linewidth]{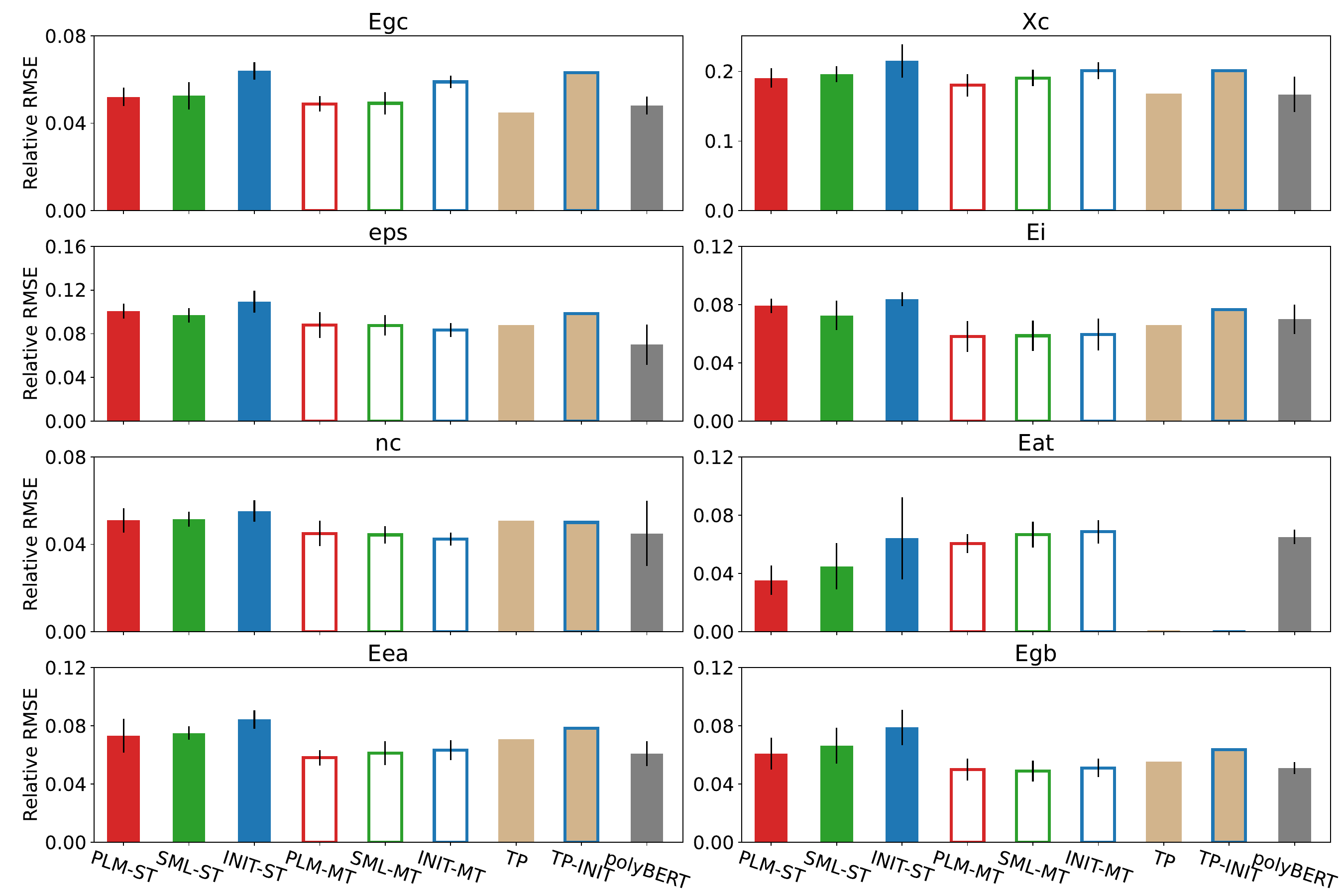}
\caption{Relative RMSE prediction error of DFT properties with BERT tokenizer. Error bars show the STD from the 5-fold cross-validation. Results of TransPolymer \cite{xu2023transpolymer} (no STD reported, hence no error bars plotted), where TP and TP-INIT represent the models with and without pre-training, respectively, and polyBERT \cite{kuenneth_polybert_2023} are also shown for reference.}
\label{fig:bar-RelativeRMSE-BERT}
\end{figure}

\begin{figure}[ht]
\centering
\includegraphics[width=0.8\linewidth]{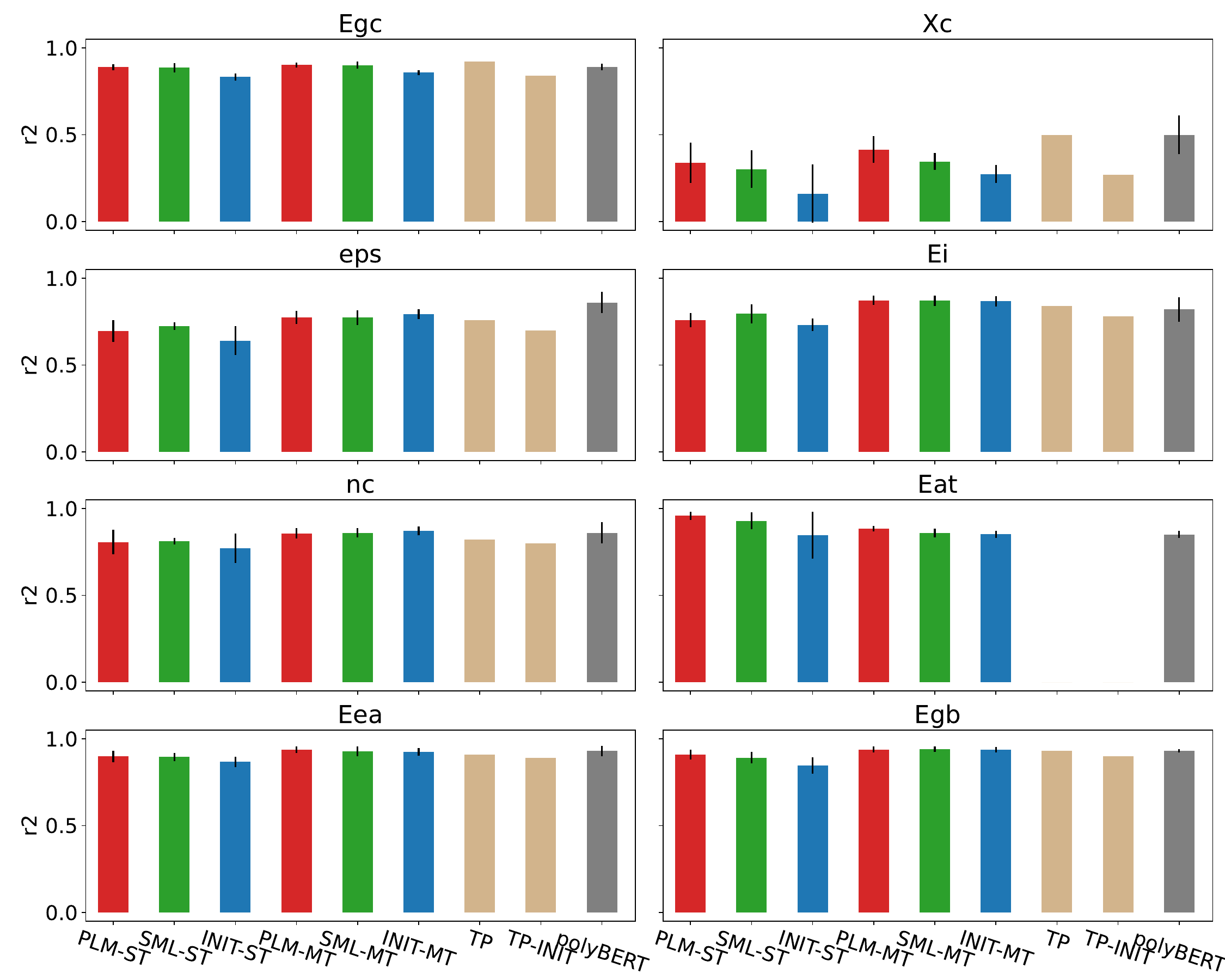}
\caption{Relative R$^2$ prediction accuracy of DFT properties with BERT tokenizer. Error bars show the STD from the 5-fold cross-validation. Results of TransPolymer \cite{xu2023transpolymer} (no STD reported, hence no error bars plotted), where TP and TP-INIT represent the models with and without pre-training, respectively, and polyBERT \cite{kuenneth_polybert_2023} are also shown for reference.}
\label{fig:bar-R2-BERT}
\end{figure}

\begin{figure}[ht]
\centering
\includegraphics[width=0.8\linewidth]{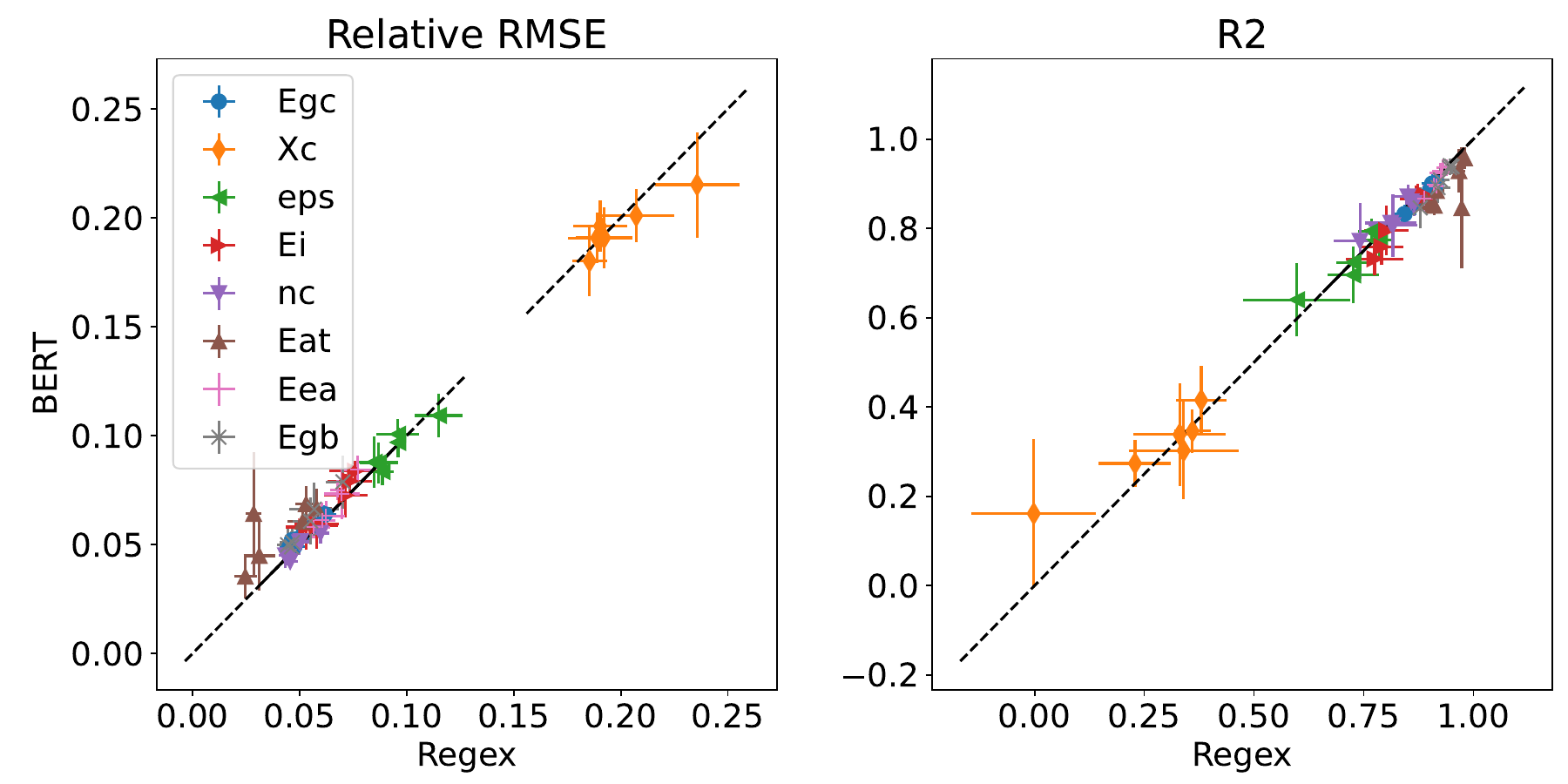}
\caption{Prediction performance: relative RMSE and R$^2$ of models with BERT tokenizer versus regex tokenizer.}
\label{fig:token-regex-BERT}
\end{figure}


\end{document}